\documentclass[preprint,12pt]{elsarticle}

\usepackage{amssymb}

\usepackage{algorithm}
\usepackage{algorithmic} 

\usepackage[english]{babel}
\addto\captionsenglish{
	
}

\usepackage{hyperref}       
\usepackage{url}            
\usepackage{amsfonts}  
\usepackage{enumerate}
\usepackage{verbatim}
\usepackage{amsmath,amsthm}
\usepackage{multirow}
\usepackage{epsfig}
\usepackage{booktabs}
\usepackage{graphicx} 
\usepackage{subcaption}
\usepackage{makecell}
\usepackage{colortbl}  
\usepackage{xcolor}
\usepackage{array}   
\usepackage{caption}
\usepackage{cleveref}
\crefname{figure}{Fig.}{Figs.}
\Crefname{figure}{Fig.}{Figs.}
\crefname{table}{Table}{Tables}
\Crefname{table}{Table}{Tables}

\usepackage{titlesec}
\usepackage{bbding}

\usepackage{makecell}

\titleformat{\paragraph}
{\normalfont\normalsize\bfseries}{\theparagraph}{0.5em}{}
\titlespacing*{\paragraph}
{0pt}{3.25ex plus 1ex minus .2ex}{1.5ex plus .2ex}

\journal{Nuclear Physics B}

\begin{document}

\begin{frontmatter}



\title{ISSTAD: Incremental Self-Supervised Learning Based on Transformer for Anomaly Detection and Localization}


\author[]{Wenping Jin \corref{cor1} \fnref{a1}}
\ead{jinwenping@stu.xjtu.edu.cn}
\cortext[cor1]{Corresponding author}

\author[]{Fei Guo \fnref{a1}}
\ead{co.fly@stu.xjtu.edu.cn}

\author[]{Li Zhu \fnref{a1}}
\ead{zhuli@xjtu.edu.cn}

\affiliation[a1]{organization={School of Software Engineering,Xi'an Jiaotong University},
	addressline={28 Xianning West Road}, 
	city={Xi'an},
	postcode={710049}, 
	state={Shaanxi},
	country={China}}

\begin{abstract}
In the realm of machine learning, the study of anomaly detection and localization within image data has gained substantial traction, particularly for practical applications such as industrial defect detection. While the majority of existing methods predominantly use Convolutional Neural Networks (CNN) as their primary network architecture, we introduce a novel approach based on the Transformer backbone network. Our method employs a two-stage incremental learning strategy. During the first stage, we train a Masked Autoencoder (MAE) model solely on normal images. In the subsequent stage, we apply pixel-level data augmentation to generate corrupted normal images and their corresponding pixel labels. Then we allow the model to learn how to repair corrupted regions and classify the status of each pixel. Ultimately, the model generates a pixel reconstruction error matrix and a pixel anomaly probability matrix. These matrices are then combined to produce an anomaly scoring matrix that effectively detects abnormal regions. When benchmarked against several state-of-the-art CNN-based methods, our approach exhibits superior performance on the MVTec AD dataset, achieving an impressive 97.6$\%$ AUC.

\end{abstract}



\begin{keyword}


Anomaly detection \sep Transformer backbone \sep Incremental learning \sep Masked Autoencoders  \sep Pixel-level Self-Supervised Learning 

\end{keyword}

\end{frontmatter}


\section{Introduction}
\label{}
Anomaly detection is a crucial area of research in machine learning, focused on identifying  anomalous data points or events within datasets\cite{1,2}. For humans, recognizing anomalies is often instinctual; even when encountering an abnormal phenomenon for the first time, it can be easily identified with sufficient knowledge of normal information. Similarly, our goal is to equip machine learning models with anomaly detection capabilities. Specifically, we explore anomaly detection methods that involve training models solely on normal data and then distinguishing between normal and abnormal data during the inference phase. Early anomaly detection methods were primarily designed for low-dimensional, small-sample data \cite{3,4,5}. However, with the emergence of deep learning and other advanced technologies, anomaly detection has extended to high-dimensional and complex data processing, such as image anomaly detection\cite{6}. 
\par
Most image anomaly detection methods based on the entire image's features. Nevertheless, in practice, anomalies in images are often localized, such as hazardous items in security inspection images \cite{7}, localized lesions in medical images \cite{8}, or localized defects in industrial product images \cite{9}. These local anomalies do not necessarily impact the overall feature information of the image. To address this problem, a widely used method involves segmenting the image into patches and identifying the abnormal state of each patch. For instance, P-SVDD\cite{28} and GP\cite{47} divide an image into multiple patches and design anomaly detection tasks for each patch, enabling the model to detect local anomalies. However, patch-level anomaly detection methods depend on the division of patches, which might not be sufficiently fine-grained. Another popular method is based on reconstruction models using denoising autoencoders. Approaches like \cite{18,19} have shown that occluding parts of an image and learning to reconstruct the occluded area can enhance the model's ability to identify local anomalies. This is because, during the process of reconstructing the occluded area, the model must learn the correlation information between different regions. Although these methods also use a patch-level approach, the patch-level reconstruction model has a significant drawback, as it may focus more on the correlation between patches and ignore some high-frequency information in the image.
\par
All the methods mentioned above use CNN as the backbone networks. Kaiming He recently proposed a patch-level occlusion and reconstruction model called MAE \cite{20}, which is based on the ViT \cite{21} autoencoder, differing from most CNN-based autoencoders. The anomaly detection method based on MAE, MemMC-MAE \cite{23}, has been shown to significantly improve the model's ability to detect local anomalies in images, thanks to the powerful context understanding ability of the Transformer \cite{22}. However, MemMC-MAE, being a patch-level reconstruction method, shares the same shortcomings as other patch-level reconstruction methods.
\par
\textbf{In our approach}. The Transformer is utilized as the backbone, integrating a Vision Transformer (ViT) as the encoder and a decoder composed of multiple Transformer blocks containing multi-headed self-attention (MSA), multilayer perceptron (MLP) layers and Layer normalization(LN). We also incorporate a reconstruction head (RCH) at the end of the decoder for image reconstruction, as well as a pixel classifier head (PCH) that extracts feature information from each level of the decoder for pixel classification. To extend our method from patch-level to pixel-level, we employ the concept of incremental learning \cite{24} and train the model in two stages, with each stage using a distinct method and input data:
\par
\textbf{First stage}. We follow the same method as MAE for training: dividing the input image into multiple patches, masking 75$\%$ of them, and then reconstructing these masked patches. During this stage, the model focuses on using normal images as inputs and only activates the reconstruction head. We update both the encoder and decoder parameters to achieve two objectives: allowing the model to initially learn the distributional characteristics of normal images and enhancing the encoder's feature extraction capabilities.
\par
\textbf{Second stage}. The model is trained by exclusively updating the decoder's parameters, which enables it to identify normal and pseudo-anomaly images, segment anomalous regions, and restore pseudo-anomaly images to their original state. To accomplish this, we develop an innovative pixel-level data augmentation technique that introduces random corruptions to normal images, with the size, location, and number of corrupted regions determined randomly. Concurrently, a label matrix is generated, with its elements signifying whether the corresponding pixel is corrupted. Throughout the training process, the model is expected to learn how to repair the corrupted areas while employing a pixel-level classifier to categorize each pixel. This approach ensures that the classifier's output aligns closely with the label matrix. As a consequence, the model can obtain both the pixel-level reconstruction error matrix and the pixel-level abnormal probability matrix. These two matrices are subsequently merged to produce an abnormal score matrix during the inference stage.
\par
Our experimental results in Section \ref{Ablation experiments} demonstrate that the two-stage training method significantly improves the model's performance. Furthermore, the fusion model used in the second stage effectively leverages the advantages of both reconstruction and classification models, leading to optimal results. We also compare our method with other state-of-the-art anomaly detection methods in MVTec AD datasets, achieving the best performance in both anomaly detection and anomaly localization.
\par
In summary, our contributions are as follows:
\par
\begin{itemize}
	\item We employ a Transformer as the backbone network for our method, outperforming state-of-the-art anomaly detection methods that use CNN as their backbone network in both anomaly detection and localization on the MVTec AD datasets.
	\item A two-stage incremental learning training method for anomaly detection is developed, leading to a significant improvement in the model's performance for both anomaly detection and localization tasks.
	\item We create a novel pixel-level data enhancement method, and building on this, design an innovative self-supervised learning method for anomaly detection that enhances the model's ability to identify local anomalies.
	\item A fusion model for anomaly detection is introduced, which effectively combines the advantages of both reconstruction-based and pixel classification-based approaches to enhance performance.
\end{itemize}
\par
Next, in Section 2, we will introduce some research related to our method. Section 3 will provide a detailed description of our method. Section 4 will present the comparison results between our method and related methods, and demonstrate the effectiveness of our method through ablation experiments. Finally, in Section 5, we will summarize the entire paper.
\par
\section{Related Work}
\label{}
\subsection{Classifier-Based Anomaly Detection}
\label{}
Classifier-based anomaly detection can be divided into two types: one-classifier based methods and data augmented classifier based methods:
\par
One-classifier based methods map positive samples to a hyperplane and optimize parameters to maximize (or minimize) the distance between the samples and the center point. This allows for determining whether a sample is anomalous based on its distance from the center point. Early one-class anomaly detection methods, such as OCSVM\cite{5} and SVDD\cite{26}, were based on support vector machines\cite{25} and demonstrated strong performance in handling small sample, low-dimensional data. However, these methods struggle with high-dimensional complex image datasets due to computational resource limitations. DeepSVDD\cite{27}, a deep learning extension of SVDD, reduces computational complexity and increases the method's generality by using deep neural networks to reduce high-dimensional data to low-dimensional space. Nevertheless, DeepSVDD is less effective in detecting small defects in image anomaly detection, such as medical images and industrial inspections, as it prioritizes overall semantic information over local defects. To address this issue, Jihun Yi et al. proposed the P-SVDD\cite{28} method. This approach divides an image into multiple patches and performs anomaly detection on each patch, improving the accuracy of both anomaly detection and localization. However, P-SVDD requires manual selection of patch size, which may lead to inaccurate anomaly detection results if not appropriately selected. Our method takes inspiration from one-class anomaly detection methods and requires the model to output a zero matrix when the input is a normal sample. In contrast to P-SVDD's patch-level classifier, our classifier is more refined and classifies each pixel.
\par
Typical data augmented classifier based methods rely on image set geometric transformations\cite{11,12,13,14} such as translation, symmetry, and rotation operations on normal images to train a classifier to recognize these operations. Anomalous images are considered those for which the classifier cannot correctly identify the transformations. These methods have achieved good performance in anomaly detection on datasets such as CIFAR10\cite{44} and F-MNIST\cite{45}. However, they mainly focus on the overall structural information of images. For industrial detection datasets like MVTec AD\cite{46}, where local defects may not affect the overall structural information of the image, these methods may lead to misjudgment. Therefore, Chun-Liang Li et al. proposed a new data augmentation method called CutPaste\cite{48} to enable the classifier model to focus on local anomalies. This method modifies local areas of the image and then identifies the original and modified image through the classifier. However, CutPaste uses connected corrupted areas to simulate abnormalities, while in reality, surface defects are often non-connected areas. Moreover, CutPaste solely relies on whole-image classification loss in training, without designing a more refined pixel-level classification loss, which is not conducive to identifying small abnormal regions.
\par 
\subsection{Reconstruction-Based Anomaly Detection}
\label{}
Reconstruction-based anomaly detection methods train reconstruction models using normal samples and detect anomalies by measuring the reconstruction error \cite{15}. Generally, it is assumed that the reconstruction error for anomalous samples will be higher than that of normal samples. These methods are typically based on autoencoders. Since the application of autoencoders to anomaly detection by DAE\cite{15}, researchers have proposed various methods to improve their performance in anomaly detection: \cite{33,34,35} introduce GAN's adversarial classifier to enhance the authenticity of reconstructed images. MeanAE\cite{36} and MeanSTC\cite{37} incorporate memory modules to capture distribution information of normal data, thereby enhancing the model's robustness. Metaformer (MF)\cite{39} leverages meta-learning\cite{40} to compensate for the model adaptability of reconstruction-based methods. GP\cite{47} detects abnormal regions in the image by comparing global and local features of the image to strengthen the model's ability to detect local anomalies. UTRAD\cite{53} proposes a U-shaped Transformer structure to reconstruct the multi-scale features extracted by CNN, thereby improving the ability of the reconstruction model to identify multi-scale anomalies.
The methods mentioned above predominantly employ CNN as their backbone networks. Nevertheless, since the introduction of the MAE\cite{20} method, autoencoders based on pure Transformer\cite{22} have gained attention and have been applied to a variety of downstream tasks\cite{41,42,43}. MemMC-MAE\cite{23} was the first to use an MAE-based autoencoder for anomaly detection tasks. On the other hand, as MemMC-MAE utilizes the same training method as MAE, it tends to prioritize the correlation between patches, which may lead to the potential neglect of high-frequency information in the image.
\par
\subsection{Discussion}
\label{}
The methods mentioned above, whether based on reconstruction or classification, generate supervisory signals from unlabeled data for training purposes. As such, they can be classified as self-supervised learning approaches. Our method also falls under the self-supervised learning category, as it integrates both reconstruction and classification components. When compared to other anomaly detection methods within the self-supervised learning domain, our approach exhibits two distinguishing features. First, it implements a two-stage incremental learning method, with the training mode and input data varying between stages. Second, it employs a pure Transformer as the backbone network. To evaluate the effectiveness of our approach, we developed eight anomaly detection models, each using a pure Transformer as their backbone network. These models are either substructures of our own design or alternatives that do not incorporate the two-stage incremental learning training. For more detailed information and experimental results, please refer to Section \ref{Ablation experiments}.
\par
\section{Method}
\label{}
\subsection{Overview of Model Architecture and Training Processing}
\label{}
\begin{figure}[!htbp]
	\centering
	\includegraphics[scale=0.4]{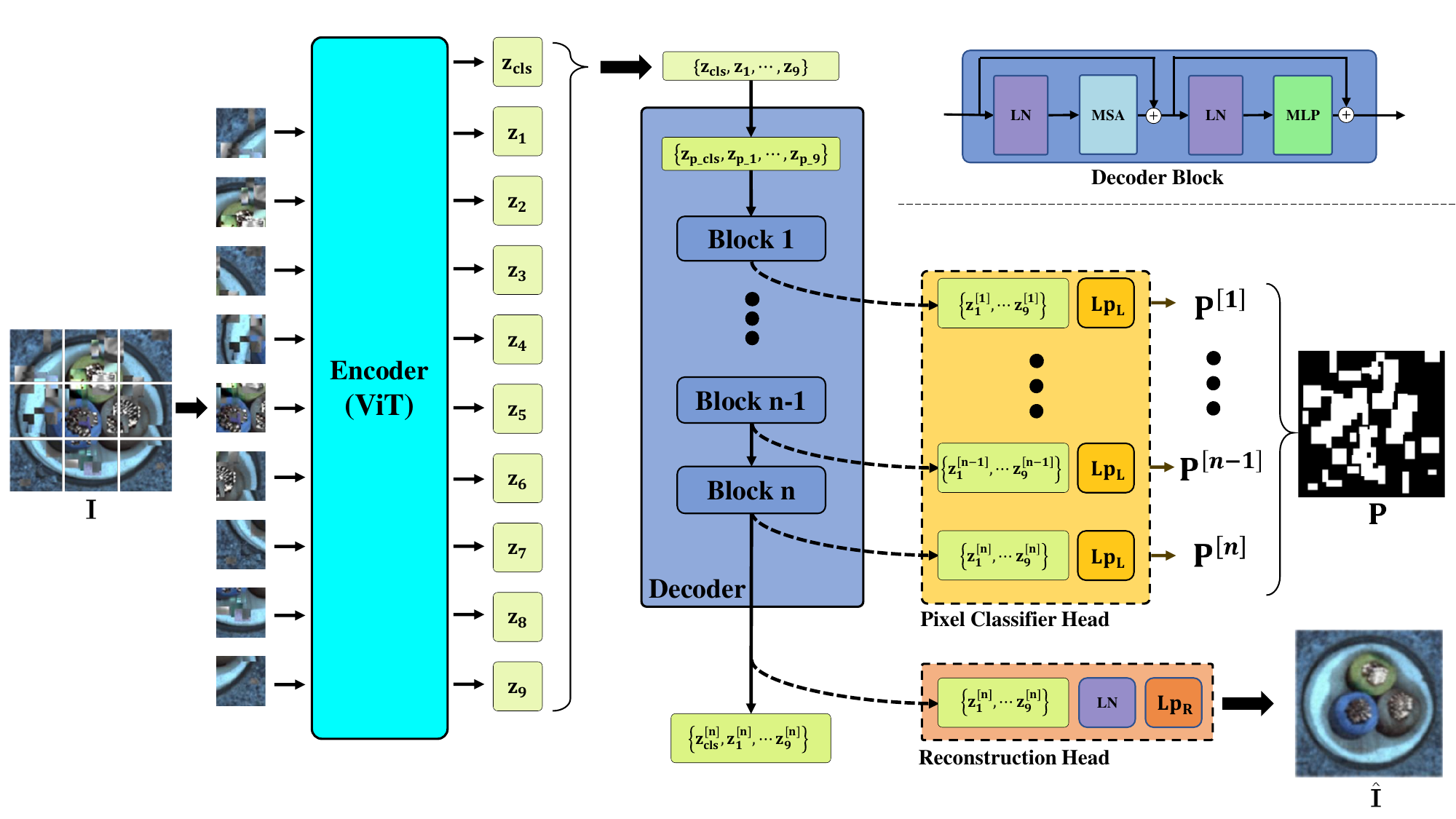}
	\caption{Model Architecture. The model consists of an encoder, decoder, reconstruction head (RCH), and pixel classifier head (PCH). The encoder employs a standard Vision Transformer (ViT), while the decoder is composed of multiple blocks. Each decoder block contains multi-headed self-attention (MSA) and multilayer perceptron (MLP) layers. Layer normalization (LN) and residual connections are also integrated within the blocks, as illustrated in the upper right corner. The reconstruction head processes output tokens from the decoder's final block, excluding the CLS token, through layer normalization and linear projection to reconstruct the image. The pixel classifier head, on the other hand, takes input from the feature tokens of all decoder blocks (excluding the CLS token) and generates a series of pixel classification matrices. These matrices are subsequently fused to form a single pixel classification matrix, which incorporates both high-level and low-level feature information.}
	\label{fig:6}
\end{figure}

In this paper, we propose a novel anomaly detection and localization model utilizing a Transformer-based backbone architecture, as illustrated in \Cref{fig:6}. The model consists of an encoder and a decoder, and is additionally equipped with a reconstruction head (RCH) for image reconstruction and a pixel classifier head (PCH) for pixel classification. The details of these components are as follows:
\par
\textbf{Encoder}: We employ a standard Vision Transformer (ViT) as the encoder. The encoder output includes a CLS token and a series of feature tokens. The CLS token aims to capture global image information, while the remaining feature tokens represent local and global information for different patches.
\par
\textbf{Decoder}: The decoder consists of several Transformer blocks, each containing multi-headed self-attention (MSA) and multilayer perceptron (MLP) layers. Layer normalization (LN) is applied before each layer, and residual connections are introduced after each layer, as illustrated in the upper right corner of \Cref{fig:6}. The Transformer block in the decoder shares the same architecture as in ViT, but serves a contrasting function, similar to deconvolution in CNNs. However, unlike deconvolution in CNNs, the Transformer block doesn't require architectural changes; it can serve different purposes by controlling information flow. The encoder output tokens, which undergo the decoder's positional embeddings and linear projection (like in ViT), serve as input to the decoder. We retain the CLS token from the encoder output but exclude it from both the reconstruction head and pixel classifier head.
\par
\textbf{Reconstruction Head}: The reconstruction head consists of a layer normalization layer and a linear projection layer. The input to the reconstruction head is the output tokens from the decoder's final block, excluding the CLS token. These tokens pass through the layer normalization layer and are ultimately transformed into a reconstructed image via the linear projection layer.
\par
\textbf{Pixel Classifier Head}: The pixel classifier head takes input from the feature tokens, excluding the CLS token, of all decoder blocks. Each block's output undergoes a linear projection and is reshaped into a pixel classification probability matrix, with each element representing the corresponding pixel's classification probability. This process generates a series of pixel classification matrices, which are then fused to form a single matrix. The rationale behind this approach is that the decoder, ending with the reconstruction head, processes high-level features from the encoder output to the reconstructed image. As a result, feature tokens from each block's output gradually transition from high-level semantic features to low-level texture features, while the global information within these tokens decreases. For anomaly detection, it is crucial to combine semantic and texture information, as well as domain and global information, to accurately identify anomalous image regions.
\par
\begin{figure}[!htbp]
	\centering
	\includegraphics[scale=0.4]{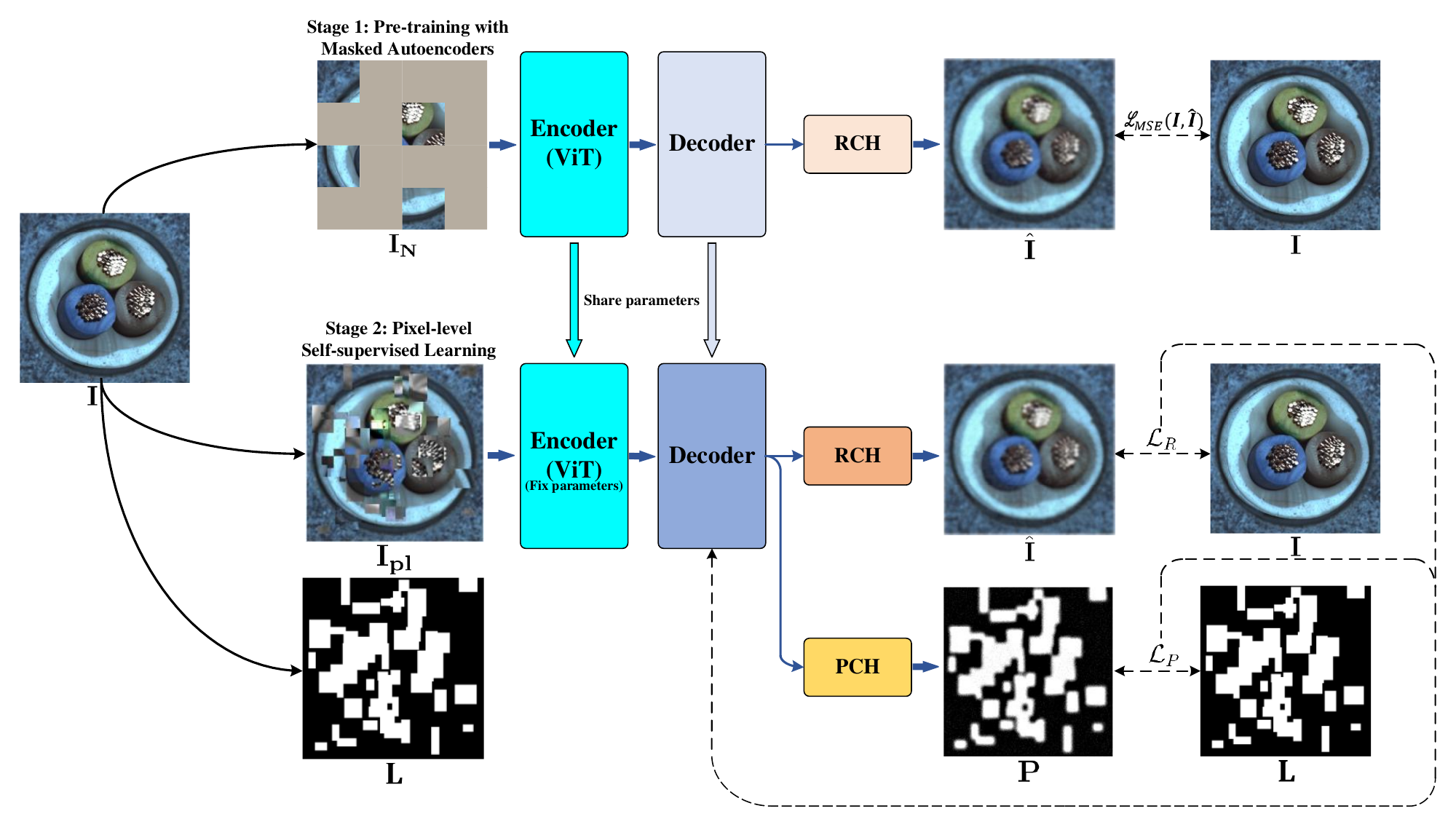}
	\caption{Overview of our training process. The process comprises two stages: Stage 1 trains the model using the same random masking and reconstruction method as employed by MAE. Stage 2 focuses on learning to repair corrupted regions and segment normal and corrupted regions within the image using a pixel-level classifier. In Stage 2, the model's initial parameters are derived from Stage 1. During training in Stage 2, we maintain fixed encoder parameters while updating only the decoder parameters for optimal performance.}
	\label{fig:1}
\end{figure}

\textbf{Model Training Process}: As depicted in \Cref{fig:1}, the model training process consists of two stages: (1) training the model utilizing the random masking and reconstruction method, similar to MAE, and (2) learning to repair corrupted regions and segment normal and corrupted regions in the image via a pixel-level classifier. The detailed implementation process is described below.
\par
\par
\subsection{Stage 1: Pre-training with Masked Autoencoders}
\label{}
In this stage, we utilized the MAE method, as illustrated in \Cref{fig:1}. The original image was divided into multiple patches, with 75$\%$ randomly masked. The encoder applies ViT on visible, unmasked patches, embedding them using linear projection and positional embeddings. By operating on a small subset (25$\%$) of the full set and excluding masked patches and mask tokens, the encoder enables efficient training of large models. The decoder processes the full token set (encoded visible patches, mask tokens), using mask tokens as shared, learned vectors for missing patches and positional embeddings for location information, then processing with Transformer blocks. The model's objective is to reconstruct the entire image using only 25$\%$ of the original information, activating only the Reconstruction Head (${\mathbf{R}\mathbf{C}\mathbf{H}}$) for this purpose.
\par
We define $\mathbf{I}$ as the set of all patches in the original image, $\mathbf{I}_{\mathbf{M}}$ as the set of masked patches in the original image, and $\mathbf{I}_{\mathbf{N}}$ as the set of unmasked patches. Thus, $\mathbf{I} = \mathbf{I}_{\mathbf{M}} \cup \mathbf{I}_{\mathbf{N}}$. We also define $\hat{\mathbf{I}}$ as the set of all patches in the reconstructed image, ${\hat{\mathbf{I}}}_{\mathbf{M}}$ as the set of masked patches in the reconstructed image, and ${\hat{\mathbf{I}}}_{\mathbf{N}}$ as the set of unmasked patches. Consequently, $\hat{\mathbf{I}} = {\hat{\mathbf{I}}}_{\mathbf{M}} \cup {\hat{\mathbf{I}}}_{\mathbf{N}}$.

We denote the encoder as ${\mathbf{E}\mathbf{n}\mathbf{c}}$, the decoder as ${\mathbf{D}\mathbf{e}\mathbf{c}}$, and the Reconstruction Head as ${\mathbf{R}\mathbf{C}\mathbf{H}}$. $\left\{ {{\mathbf{Z}_{cls}^{\lbrack n\rbrack},\mathbf{Z}}_{1}^{\lbrack n\rbrack},\cdots\mathbf{Z}_{k}^{\lbrack n\rbrack}} \right\}$ is the output of decoder, k is the number of patches . Among them, ${\mathbf{Z}_{cls}^{\lbrack n\rbrack}}$ is not used as the input of ${\mathbf{R}\mathbf{C}\mathbf{H}}$.
\par
\begin{equation} \label{eqZ}
	\left\{ {{\mathbf{Z}_{cls}^{\lbrack n\rbrack},\mathbf{Z}}_{1}^{\lbrack n\rbrack},\cdots\mathbf{Z}_{k}^{\lbrack n\rbrack}} \right\}~ = {\mathbf{D}\mathbf{e}\mathbf{c}\left\lbrack {\mathbf{E}\mathbf{n}\mathbf{c}\left( \mathbf{I}_{\mathbf{N}}  \right)} \right\rbrack},
\end{equation}
\par
\begin{equation} \label{eq1}
	\hat{\mathbf{I}}~ = \mathbf{R}\mathbf{C}\mathbf{H}\left( \left\{ {\mathbf{Z}_{1}^{\lbrack n\rbrack},\cdots\mathbf{Z}_{k}^{\lbrack n\rbrack}} \right\} \right).
\end{equation}
\par
The loss function is the same as MAE, which calculates the mean squared error (MSE) between the reconstructed image and the original image. To reduce computation, only the reconstruction loss of masked image blocks is calculated here:
\par
\begin{equation} \label{eq2}
	\mathcal{L}_{MSE}\left( \mathbf{I},\hat{\mathbf{I}} \right)\left. = \right\|\mathbf{I}_{\mathbf{M}} - {\hat{\mathbf{I}}}_{\mathbf{M}}\|_{2}^{2}.
\end{equation}
\par
By doing so, we assumed that the model learned the distribution characteristics of normal images, which is crucial for anomaly detection tasks. These tasks require learning the distribution characteristics of normal images and parameterizing those characteristics. If the parameter of a particular image's distribution characteristics differs from the normal parameter, then the image is judged to be an anomaly. In Section \ref{Ablation experiments}, we will show that even the most basic MAE method performs well in anomaly detection and localization when an appropriate distribution parameterization method is selected.
\par
\subsection{Stage 2: Pixel-level Self-supervised Learning}
\label{}
\par
\subsubsection{Data Augmentation}
\label{}
\par
\begin{figure}[!htbp]
	\centering
	\includegraphics[scale=0.6]{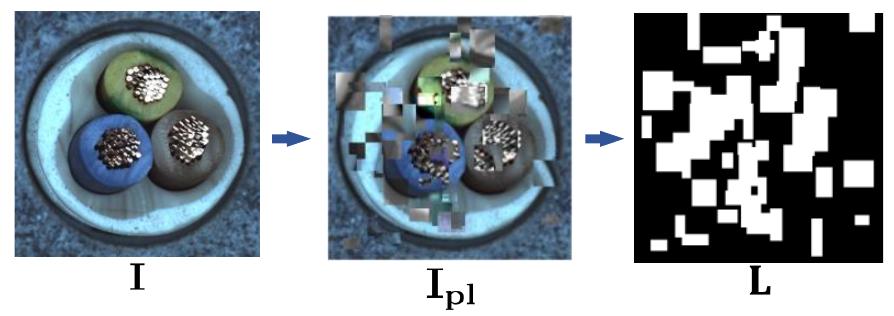}
	\caption{Overview of our data augmentation method. We randomly apply Gaussian noise, alter color channels, flip, and perform other operations on randomly sized and numbered image blocks in image $\mathbf{I}$ to generate the corrupted image $\mathbf{I}_{\mathbf{p}\mathbf{l}}$. Simultaneously, we produce a pixel-level label matrix M. Each element in matrix $\mathbf{L}$ corresponds to a pixel in the image; if the pixel remains unchanged, the corresponding element in matrix $\mathbf{L}$ is 0; otherwise, the corresponding element in matrix $\mathbf{L}$ is 1.}
	\label{fig:2}
\end{figure}

We propose a novel pixel-level data augmentation method, as illustrated in \Cref{fig:2}. We randomly apply Gaussian noise, alter color channels, flip, and perform other operations on randomly sized and numbered image blocks in image $\mathbf{{I}}\in \mathbb{R}^{H \times W \times C}$ to generate the corrupted image $\mathbf{I}_{\mathbf{p}\mathbf{l}} \in \mathbb{R}^{H \times W \times C}$. Simultaneously, we produce a pixel-level label matrix $\mathbf{{L}}\in \mathbb{R}^{H \times W}$. Each element in matrix $\mathbf{L}$ corresponds to a pixel in the image; if the pixel remains unchanged, the corresponding element in matrix $\mathbf{L}$ is 0; otherwise, the corresponding element in matrix $\mathbf{L}$ is 1. Matrix $\mathbf{L}$ can be utilized as a pixel classification label. During training, not all model inputs are corrupted images; instead, a specific proportion of normal images and corrupted images are used. For normal images that have not been corrupted, all elements in the $\mathbf{L}$ matrix are $\boldsymbol{0}$.
\par
\par
\subsubsection{Calculating the Reconstructed Image and Pixel Classification Matrix}
\label{}
During the training phase, the model takes either original or corrupted images as input. As discussed in the previous section, normal images are randomly corrupted to a certain extent during the image augmentation stage. This means that the model's input can be either $\mathbf{I} \in \mathbb{R}^{H \times W \times C}$ (original) or $\mathbf{I}_{\mathbf{p}\mathbf{l}} \in \mathbb{R}^{H \times W \times C}$ (corrupted). Both the reconstruction head(${\mathbf{R}\mathbf{C}\mathbf{H}}$) and the pixel classifier head(${\mathbf{P}\mathbf{C}\mathbf{H}}$) are activated in this stage.
\par
The output tokens of the encoder are represented as $\left\{ {\mathbf{z}_{\mathbf{c}\mathbf{l}\mathbf{s}},\mathbf{z}_{1},\mathbf{~}\cdots,\mathbf{z}_{k}} \right\}$. In the head of decoder, these tokens should be undergoing positional embedding and linear projection from the decoder, then these tokens can be represented as $\left\{ {\mathbf{z}_{\mathbf{p}\_\mathbf{c}\mathbf{l}\mathbf{s}},\mathbf{z}_{\mathbf{p}\_ 1},\mathbf{~}\cdots,\mathbf{z}_{\mathbf{p}\_\mathbf{k}}} \right\}$, we denote this set by ${\mathbf{Z}}^{[0]}$. Then the output of the $\ell$th block is ${\mathbf{Z}}^{[\ell]}$, which is calculated by the following:
\par
\begin{equation}
	\begin{aligned}
		& \hat{\mathbf{Z}}^{[\ell]}=\operatorname{MSA}\left(\operatorname{LN}\left(\mathbf{Z}^{[\ell-1]}\right)\right)+\mathbf{Z}^{[\ell-1]}, \quad \ell=1 \ldots n,\\
		& \mathbf{Z}^{[\ell]}=\operatorname{MLP}\left(\operatorname{LN}\left(\hat{\mathbf{Z}}^{[\ell]}\right)\right)+\hat{\mathbf{Z}}^{[\ell]}, \quad \ell=1 \ldots n.\\
	\end{aligned}
\end{equation}
\par
Among them, LN represents layer normalization, MSA stands for multi-headed self-attention, and MLP refers to multilayer perceptron layers. The variable '$n$' denotes the number of decoder transformer blocks. The set ${\mathbf{Z}}^{[\ell]}$ consists of tokens for $\left\{{{\mathbf{z}_{\mathbf{c}\mathbf{l}\mathbf{s}}^{\lbrack\mathbf{\ell}\rbrack},\mathbf{z}}_{1}^{\lbrack\mathbf{\ell}\rbrack},\cdots\mathbf{z}_{k}^{\lbrack\mathbf{\ell}\rbrack}} \right\}$. This set is used as the input for ${\mathbf{P}\mathbf{C}\mathbf{H}}$, excluding the token ${\mathbf{z}_{cls}^{\lbrack n\rbrack}}$. Thus, the $\ell$th pixel classification matrix $\mathbf{P}^{\lbrack \ell\rbrack} \in [0, 1]^{H \times W}$ has the same height and width dimensions as the input image. The elements in the matrix represent the probability of the corresponding pixel in the input image being abnormal. The calculation is as follows:
\par
\begin{equation} \label{eq4} 
	\mathbf{P}^{\lbrack \ell\rbrack} = {\mathbf{L}\mathbf{p}}_{\mathbf{L}}\left( \left\{ {\mathbf{z}_{1}^{\lbrack \ell\rbrack},\cdots\mathbf{z}_{k}^{\lbrack \ell\rbrack}} \right\} \right).
\end{equation}
\par
As result, we can get a set of pixel classification matrices $\left\{ {\mathbf{P}^{\lbrack 1\rbrack},\cdots\mathbf{P}^{\lbrack n\rbrack}} \right\}$.
\par
${\mathbf{R}\mathbf{C}\mathbf{H}}$ is responsible for image reconstruction, with the aim of ensuring that the reconstructed image $\mathbf{\hat{I}}\in \mathbb{R}^{H \times W \times C}$ closely matches the original image $\mathbf{I}$, regardless of whether it has been corrupted or not.$\left\{{{\mathbf{z}_{\mathbf{c}\mathbf{l}\mathbf{s}}^{\lbrack\mathbf{n}\rbrack},\mathbf{z}}_{1}^{\lbrack\mathbf{n}\rbrack},\cdots\mathbf{z}_{k}^{\lbrack\mathbf{n}\rbrack}} \right\}$ is the output of decoder, among them, ${\mathbf{z}_{cls}^{\lbrack n\rbrack}}$ is not used as the input of ${\mathbf{R}\mathbf{C}\mathbf{H}}$.
\par
\begin{equation} \label{eq3a}
	\left\{ {{\mathbf{z}_{cls}^{\lbrack n\rbrack},\mathbf{z}}_{1}^{\lbrack n\rbrack},\cdots\mathbf{z}_{k}^{\lbrack n\rbrack}} \right\}~ = {\mathbf{D}\mathbf{e}\mathbf{c}\left\lbrack {\mathbf{E}\mathbf{n}\mathbf{c}\left(  \mathbf{I} \text{ or }  \mathbf{I}_{\mathbf{p}\mathbf{l}}   \right)} \right\rbrack},
\end{equation}
\par
\begin{equation} \label{eq3b}
	\hat{\mathbf{I}}~ = \mathbf{R}\mathbf{C}\mathbf{H}\left( \left\{ {\mathbf{z}_{1}^{\lbrack n\rbrack},\cdots\mathbf{z}_{k}^{\lbrack n\rbrack}} \right\} \right).
\end{equation}
\par
During the training initialization phase, the initial parameters of the encoder and decoder are derived from first stage. Throughout the second training stage, we maintain fixed encoder parameters and update only the decoder parameters.
\par
\subsubsection{Loss Functions}
\label{}
In this stage, we focus on two training tasks: the reconstruction task, where the output reconstruction image $\mathbf{\hat{I}}$ should closely resemble the original image $\mathbf{I}$, and the pixel classification task, where the pixel classifier should output the probability of each pixel being corrupted.
\par
For the reconstruction task, we first consider the mean square error (MSE) loss. In contrast to the pretraining loss in Section 3.2, we use the entire image for the MSE loss:
\par
\begin{equation} \label{eq5}
	\mathcal{L}_{MSE}\left( \mathbf{I},\hat{\mathbf{I}} \right)\left. = \right\|\mathbf{I} - \hat{\mathbf{I}}\|_{2}^{2}.
\end{equation}
\par
However, if only MSE is used as the loss function, the model may produce overly smooth reconstructed images, causing some structured information to be overlooked. To address this issue, we introduce the SSIM \cite{49} loss function as a supplement. The SSIM loss function considers the error between the structured information of the original image and the reconstructed image, resulting in improved reconstruction results. The SSIM loss is defined as follows:
\par
\begin{equation} \label{eq6}
	\mathcal{L}_{SSIM}\left( {\mathbf{I},\hat{\mathbf{I}}} \right) = \frac{1}{H \times W}{\sum_{i = 1}^{H}{\sum_{j = 1}^{W}1}} - {\mathit{SSIM}(}\mathbf{I},\hat{\mathbf{I}})_{({i,j})}.
\end{equation}
\par
The SSIM value is the Structural Similarity Index calculated between two image blocks $\mathbf{I}$ and $\mathbf{\hat{I}}$, centered at (i,j). 
\par
In the pixel classification task, we assess the discrepancy between each pixel classification matrix $\mathbf{P}^{\lbrack \ell\rbrack} \in [0, 1]^{H \times W}$, $\ell=1 \ldots n$, and the classification labels $\mathbf{L}$ using the cross-entropy loss function, a prevalent method in image segmentation\cite{51}. The cross-entropy loss function is defined as:
\par
\begin{equation} \label{eq7}
	\begin{aligned}
	\mathcal{L}_{CE}^{\lbrack \ell\rbrack}\left( {\mathbf{L},\mathbf{P}^{\lbrack\mathbf{\ell}\rbrack}} \right)
	=& - \frac{1}{H \times W}{\sum_{i = 1}^{H}{\sum_{j = 1}^{W}\omega}}*l_{({i,j})}*{\mathit{\log}\left( p_{({i,j})}^{\lbrack \ell\rbrack} \right)}\\
	& + \left( {1 - l_{({i,j})}} \right)*{\mathit{\log}\left( {1 - p_{({i,j})}^{\lbrack \ell\rbrack}} \right)}, \quad \ell=1 \ldots n.\\
	\end{aligned}
\end{equation}
\par

Here, $l_{(i,j)}$ and $p_{(i,j)}^{\lbrack \ell\rbrack}$ 
represent elements in matrices $\mathbf{L}$ and $\mathbf{P}^{\lbrack \ell\rbrack}$, respectively. We define $\mathcal{L}_{CE}$ as the mean of set $\left\{ {\mathcal{L}_{CE}^{\lbrack 1\rbrack},\cdots\mathcal{L}_{CE}^{\lbrack n\rbrack}} \right\}$:
\par
\begin{equation} \label{eqlce}
\mathcal{L}_{CE} = ~\frac{1}{n}{\sum\limits_{\ell = 1}^{n}\mathcal{L}_{CE}^{\lbrack \ell\rbrack}}.
\end{equation}
In summary, the loss function during the anomaly detection training phase is defined as:
\par
\begin{equation} \label{lr}
	\begin{aligned}
\mathcal{L}_{R} = {\lambda_{0}*\mathcal{L}}_{MSE} + {\lambda_{1}*\mathcal{L}}_{SSIM}
	\end{aligned},
\end{equation}
\par
\begin{equation} \label{lm}
	\begin{aligned}
		\mathcal{L}_{P} = {\lambda_{2}*\mathcal{L}}_{CE},
	\end{aligned}
\end{equation}
\par
\begin{equation} \label{loss}
	\begin{aligned}
	\mathcal{L} = \mathcal{L}_{R} + \mathcal{L}_{P}.
	\end{aligned}
\end{equation}
\par
In this equation, $\mathcal{L}_{R}$ is the reconstruction loss, and $\mathcal{L}_{P}$ is the pixel classification loss. $\lambda_{0}$, $\lambda_{1}$, and $\lambda_{2}$ denote the coefficients of the MSE loss, SSIM loss, andpixel classification loss, respectively. These coefficients determine the weight of each loss component within the overall loss function $\mathcal{L}$.
\par
\subsection{Training Method Discussion}
\label{}
\par
We adopt the aforementioned training method for the following reasons:
\par
\begin{itemize}
	\item Our model is an autoencoder structure composed of Transformer blocks, featuring both pixel-level classification and reconstruction functions. The Transformer blocks of the encoder and decoder share the same architecture. Without controlling the flow of information during training, the distinction between the encoder and decoder may become blurred, making it difficult to optimally train the model. The information flow control in the first stage of training assigns different tasks to the encoder and decoder. The encoder's input consists of partial image patches, while the decoder's task is to reconstruct the entire image using incomplete feature tokens. In the second stage, the control method for information flow is more direct: the encoder's parameters are fixed, and only the decoder's parameters are updated.
	\par
	\item The first stage of training essentially teaches the model to learn the distribution characteristics of normal data using a masked autoencoder. The second stage of training primarily focuses on simulating and identifying abnormal states through data enhancement. Since the encoder parameters are not updated during the second stage, the encoder's feature extraction ability for the normal part of the image is stronger than for the abnormal part. This difference makes it more challenging for the decoder to repair the abnormal portion of the image. The pixel classifier head captures this difficulty from feature tokens of different levels, providing comparative information rather than directly learning abnormal part features. Omitting the first stage of training could cause the model to concentrate on learning the characteristics of simulated anomalies, making its effectiveness highly dependent on the fidelity of the simulated anomalies.
	\par
	\item During the first training stage, the decoder is tasked with reconstructing the masked patches and primarily focuses on the relationships between patches. In the second training stage, the decoder is required to restore corrupted regions, classify each pixel, and determine whether it is in a corrupted region. The range of these corrupted regions is random, potentially spanning across patches or only affecting a small part of a patch. This forces the decoder to pay closer attention to the relationships between pixels.
\end{itemize}
\par
In conclusion, this two-stage training method employs an incremental learning approach. By incorporating two distinct training stages with varying input data and forms, the model captures more detailed and comprehensive features of normal images, ultimately enhancing its anomaly detection and localization capabilities.
\subsection{Inference of Anomaly Detection and Localization}
\label{}
\begin{figure}[h!tbp]
	\centering
	\includegraphics[scale=0.6]{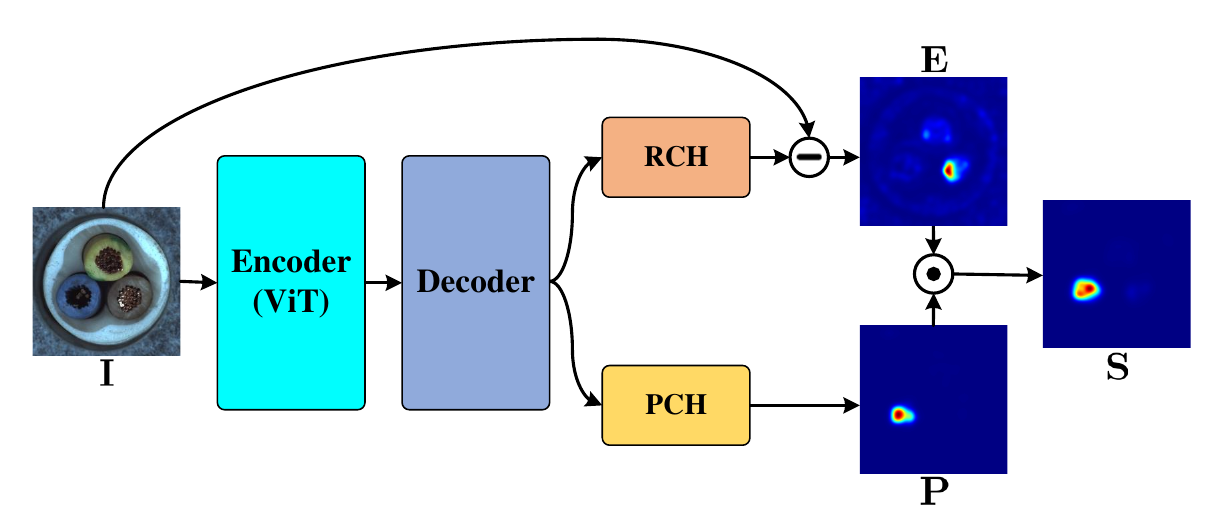}
	\caption{Overview of our inference stage method. The anomaly scoring matrix $\mathbf{S}$ is created by combining the reconstruction error matrix $\mathbf{E}$ and pixel anomaly classification probability matrix $\mathbf{P}$. }
	\label{fig:3}
\end{figure}

During the inference stage, we fuse pixel-level reconstruction error and pixel classification probability to evaluate the anomaly score for the entire image and individual pixels. The inference phase is depicted in \Cref{fig:3}.
\par
Given the input image $\mathbf{I}$, we obtain the reconstructed image $\hat{\mathbf{I}}$ by reconstruction head (\textbf{RCH}) and a set of pixel classification probability matrices $\left\{ {\mathbf{P}^{\lbrack 1\rbrack},\cdots\mathbf{P}^{\lbrack n\rbrack}} \right\}$ by pixel classifier head(\textbf{PCH}).
\par
To compute the reconstruction error matrix $\mathbf{E} \in \mathbb{R}^{H \times W}_{\geq 0}$ between the original image and the reconstructed image, we average the reconstruction errors of each channel in RGB, as follows:
\par
\begin{equation} \label{eq10}
	\mathbf{E}_{\mathbf{R}} = \left( {\mathbf{I}_{\mathbf{R}} - {\hat{\mathbf{I}}}_{\mathbf{R}}} \right)^{2}, 
\end{equation}
\par
\begin{equation} \label{eq11}
	\mathbf{E}_{\mathbf{G}} = \left( {\mathbf{I}_{\mathbf{G}} - {\hat{\mathbf{I}}}_{\mathbf{G}}} \right)^{2}, 
\end{equation}
\par
\begin{equation} \label{eq12}
	\mathbf{E}_{\mathbf{B}} = \left( {\mathbf{I}_{\mathbf{B}} - {\hat{\mathbf{I}}}_{\mathbf{B}}} \right)^{2}, 
\end{equation}
\par
\begin{equation} \label{eq13}
	\mathbf{E} = \frac{\mathbf{E}_{\mathbf{R}} + \mathbf{E}_{\mathbf{G}} + \mathbf{E}_{\mathbf{B}}}{3}.
\end{equation}
\par
The pixel classification matrix $\mathbf{P}$, witch integrates pixel classification matricis with different levelsis, and is calculating as:
\begin{equation} \label{pcal}
	\mathbf{P} = ~\frac{1}{n}{\sum\limits_{\ell = 1}^{n}\mathbf{P}^{\lbrack \ell\rbrack}}.
\end{equation}
Next, to eliminate noise in the reconstruction error matrix and the pixel classification matrix, we apply a Gaussian filter to smooth $\mathbf{E}$ and $\mathbf{P}$. To create a unified anomaly scoring matrix that combines both reconstruction error and anomaly classification probability, we define it as the Hadamard product of the reconstruction error matrix $\mathbf{E}$ and the pixel anomaly classification probability matrix $\mathbf{P}$. Consequently, the anomaly score matrix can be expressed as:
\par
\begin{equation} \label{eq14}
	\mathbf{S} = \mathbf{E}~\mathbf{\odot}~\mathbf{P}.
\end{equation}
\par
The anomaly score for a pixel can be calculated using:
\par
\begin{equation} \label{eq15}
	s_{({i,j})} = e_{({i,j})}*p_{({i,j})}.
\end{equation}
\par
In this equation, $s_{({i,j})}$ denotes the anomaly score of pixel $(i,j)$. A higher score suggests a higher probability of the pixel being abnormal. The variables $e_{({i,j})}$ and $p_{({i,j})}$ represent the reconstruction error and anomaly classification probability of pixel $(i,j)$, respectively.
\par
To calculate the anomaly score for the entire image, we compute the mean of all elements in the anomaly score matrix:
\par
\begin{equation} \label{eq16}
	Anomaly~score = \frac{1}{H \times W}{\sum_{i = 1}^{H}{\sum_{j = 1}^{W}s_{({i,j})}}}.
\end{equation}
\par
Here, the variable $s_{({i,j})}$ is determined by the product of the reconstruction error $e_{({i,j})}$ of pixel $(i,j)$ and the anomaly classification probability $p_{({i,j})}$. This implies that the anomaly score of the entire image is essentially a weighted average of pixel reconstruction errors. The anomaly score matrix is also subjected to a Gaussian filter to remove noise.
\par
\section{Experiment}
\label{}
\subsection{Experimental settings}
\label{}
\subsubsection{Instruction of Datasets}
\label{}
We based our study on the MVTec AD dataset[45], which was developed by researchers at the Technical University of Munich in 2019 to address complex real-world defect detection problems. The dataset contains 5,354 images spanning 15 distinct categories, with 3,629 images for training and 1,725 images for testing. Training and validation images are all normal and defect-free, while testing images include both abnormal images, featuring various defects such as scratches, cracks, holes, and stains, and normal images.
\par
The MVTec AD dataset comprises images of different materials, including metal, wood, tape, and plastic, and showcases various surfaces and textures. This diversity makes the dataset more representative and applicable to a wide range of real-world scenarios. Additionally, the dataset includes pre-segmented defect masks, which can be used to evaluate the performance of defect detection methods.
\par
The dataset's limited size presents unique challenges for deep learning representations, as the number of training images for each category ranges from 60 to 391.
\par
\subsubsection{Evaluation Metrics and Baseline}
\label{}
We assessed our model's performance using the area under the receiver operating characteristic (ROC) curve (AUC), a popular metric for evaluating the accuracy of binary classification models in anomaly detection. The AUC allowed us to determine how effectively our model distinguished between normal and anomalous samples in the dataset. A higher AUC score indicates better model performance in correctly classifying samples, while a lower score suggests poorer performance. By employing AUC, we could objectively compare the effectiveness of our model with other state-of-the-art anomaly detection methods.
\par
We employed several state-of-the-art methods as baselines. For anomaly detection, we used OCGAN\cite{35}, MemAE\cite{36}, GANomaly\cite{35}, SCADN\cite{33}, RIAD\cite{18}, SSM\cite{19}, P-SVDD\cite{28}, MF\cite{39}, Cutpast\cite{48}, and GP\cite{47} as baselines. For anomaly localization, we utilized OCGAN\cite{35}, MemAE\cite{36}, AnoGAN\cite{52}, SCADN\cite{38}, MemSTC\cite{37}, GP\cite{47}, SSM\cite{19}, RIAD\cite{18}, P-SVDD\cite{28}, and Cutpast\cite{48} as baselines. These methods represent the primary types of anomaly detection methods currently available.
\par
\subsubsection{Training Parameter Settings}
\label{trainsets}
\textbf{Image Input}: We resize each input image to 3×224×224 and divide it into 14×14 patches, with each patch being 3×16×16, using the standard processing method for ViT.
\par
\textbf{Model Initialization}: In our experiments, we utilize the Mae-Pretrain-Vit-Large model, derived from the ViT-Large model. The encoder comprises 24 Vit-Blocks, while the decoder is designed to be lightweight, incorporating 8 decoder Transformer blocks with a feature token dimension of 512. The initial parameters are obtained from Haiming He's team, who trained the model on the ImageNet1K dataset. All experimental code is implemented using PyTorch.
\par
\textbf{Stage 1 Training Parameters}: We optimize the model using the AdamW optimizer and apply linear learning rate scaling similar to MAE: $lr = {lr}{base} \times batchsize/256$. The ${lr}{base}$ is set to 1e-3, the batch size is 20, the weight decay is 0.05, and the momentum is $\beta 1 = 0.9,~\beta 2 = 0.95$. We employ half-cycle cosine learning rate decay and train each class of data for 120 epochs.
\par
\textbf{Data Augmentation}: During the anomaly detection training phase, in each iteration, we randomly select 5/6 of all normal samples and randomly corrupt them, while keeping the remaining samples unchanged. The random corruption is achieved by selecting 1 to 100 random-sized image blocks on the image and applying various operations such as random rotation, flipping, Gaussian noise addition, and color variation to these blocks. A single region may undergo two or more image operations simultaneously.
\par
\textbf{Stage 2 Training Parameters}: We optimize the model using the AdamW optimizer with an initial learning rate of 11e-3, weight decay of 0.05, and momentum of $\beta 1 = 0.9,~\beta 2 = 0.95$. We employ cosine annealing to adjust the learning rate during training, with a cosine period of 60 epochs, a minimum learning rate of 1e-9, and a maximum learning rate of 1e-3. The batch size is set to 16, and each class of data is trained for 200 epochs. The test results are obtained from the best-performing model over the 200 epochs.
\par
\subsection{Experimental results}
\label{}
In this section, we present the quantitative and qualitative results of the proposed method and compare it with several state-of-the-art methods on the MVTec AD dataset.
\par
\subsubsection{Anomaly detection results}
\label{}
For the anomaly detection task, \Cref{tab:1} displays the corresponding results. We compared our proposed method with several state-of-the-art techniques, such as OCGAN\cite{35}, MemAE\cite{36}, GANomaly\cite{34}, SCADN\cite{38}, RIAD\cite{18}, SSM\cite{19}, P-SVDD\cite{28}, MF\cite{39}, Cutpast\cite{48}, and GP\cite{47}. As indicated in Table 1, our method achieved the highest average AUC (97.6$\%$ AUC) across all categories.
\par
\begin{table}[htbp]
	\centering
	\caption{Anomaly Detection Performance on the MVTec AD Dataset: AUC Percentages Displayed by Category. The last row shows the average score across all categories. Data for OCGAN, MemAE, GANomaly, SCADN, and SSM is sourced from \cite{19}. RIAD results are referenced from \cite{18}, P-SVDD results are taken from \cite{28}, MF results are taken from \cite{39}, Cutpast results are taken from \cite{48}, and GP results are taken from \cite{47}. The best-performing methods are highlighted in bold.}
	\resizebox{\linewidth}{!}{
	\begin{tabular}{p{2cm}<{\centering}p{2cm}<{\centering}p{2cm}<{\centering}p{2cm}<{\centering}p{2cm}<{\centering}p{2cm}<{\centering}p{2cm}<{\centering}p{2cm}<{\centering}p{2cm}<{\centering}p{2cm}<{\centering}p{2cm}<{\centering}p{2cm}<{\centering}}
		\toprule
		Category & \textbf{OCGAN}\ & \textbf{MemAE} & \textbf{GANomaly} & \textbf{SCADN} & \textbf{RIAD} & \textbf{SSM} & \textbf{P-SVDD} & \textbf{MF} & \textbf{Cutpast} & \textbf{GP} & \textbf{OURS} \\
		 & \cite{35} & \cite{36} & \cite{34} & \cite{38} & \cite{18} & \cite{19} & \cite{28} & \cite{39} & \cite{48} & \cite{47} &   \\
		\midrule
		Bottle & 59.2  & 93    & 89.2  & 95.7  & 99.9  & 99.9  & 98.6  & 99.1  & 98.2  & 96.8  & \cellcolor[rgb]{ .949,  .949,  .949}\textbf{100} \\
		Cable & 49.6  & 78.5  & 74.5  & 85.6  & 88.4  & 77.3  & 90.3  & \textbf{97.1} & 81.2  & 98    & \cellcolor[rgb]{ .949,  .949,  .949}93.6 \\
		Capsule & 71.4  & 73.5  & 73.2  & 76.5  & 99.6  & 91.4  & 76.7  & 87.5  & \textbf{98.2} & 96    & \cellcolor[rgb]{ .949,  .949,  .949}92.9 \\
		Carpet & 34.8  & 38.6  & 69.9  & 50.4  & 100   & 76.3  & 92.9  & 94    & 93.9  & \textbf{97.7} & \cellcolor[rgb]{ .949,  .949,  .949}97.1 \\
		Grid  & 85.5  & 80.5  & 70.8  & 98.3  & 83.8  & 100   & 94.6  & 85.9  & 100   & 93.2  & \cellcolor[rgb]{ .949,  .949,  .949}\textbf{100} \\
		Hazelnut & 75.3  & 76.9  & 78.5  & 83.3  & 98.7  & 91.5  & 92    & 99.4  & 98.3  & 96.2  & \cellcolor[rgb]{ .949,  .949,  .949}\textbf{100} \\
		Leather & 62.4  & 42.3  & 84.2  & 65.9  & 90.9  & 99.9  & 90.9  & 99.2  & 100   & 90.9  & \cellcolor[rgb]{ .949,  .949,  .949}\textbf{100} \\
		MetalNut & 29.5  & 65.4  & 70    & 62.4  & 98.1  & 88.7  & 94    & 96.2  & \textbf{99.9} & 96.7  & \cellcolor[rgb]{ .949,  .949,  .949}99.3 \\
		Pill  & 70.2  & 71.7  & 74.3  & 81.4  & 81.9  & 89.1  & 86.1  & 90.1  & 94.9  & \textbf{97.8} & \cellcolor[rgb]{ .949,  .949,  .949}96.6 \\
		Screw & 50.5  & 25.7  & 74.6  & 83.1  & 84.2  & 85    & 81.3  & \textbf{97.5} & 88.7  & \textbf{100} & \cellcolor[rgb]{ .949,  .949,  .949}88.3 \\
		Tile  & 80.6  & 71.8  & 79.4  & 79.2  & 83.3  & 94.4  & 97.8  & 99    & 99.3  & 88.3  & \cellcolor[rgb]{ .949,  .949,  .949}\textbf{100} \\
		Toothbrush & 59.4  & 96.7  & 65.3  & 98.1  & 88.5  & 100   & 100   & 100   & 99.4  & 96.1  & \cellcolor[rgb]{ .949,  .949,  .949}\textbf{100} \\
		Transistor & 47.7  & 79.1  & 79.2  & 86.3  & 84.5  & 91    & 91.5  & 94.4  & 96.1  & \textbf{99.9} & \cellcolor[rgb]{ .949,  .949,  .949}96.6 \\
		Wood  & 95.9  & 95.4  & 83.4  & 96.8  & 100   & 95.9  & 96.5  & 99.2  & 99.2  & 94.1  & \cellcolor[rgb]{ .949,  .949,  .949}\textbf{100} \\
		Zipper & 36.4  & 71    & 74.5  & 84.6  & 93    & 99.9  & 97.9  & 98.6  & 99.9  & 99.2  & \cellcolor[rgb]{ .949,  .949,  .949}\textbf{100} \\
		\midrule
		Mean  & 60.6  & 70.7  & 76.2  & 81.8  & 91.7  & 92    & 92.1  & 95.8  & 96.1  & 96.1  & \cellcolor[rgb]{ .949,  .949,  .949}\textbf{97.6} \\
		\bottomrule
	\end{tabular}%
	}
	\label{tab:1}%
\end{table}%
\par

\subsubsection{Anomaly Localization results}
\label{}
For the anomaly localization task, we employed  OCGAN\cite{35}, MemAE\cite{36}, AnoGAN\cite{52}, SCADN\cite{38}, MemSTC\cite{37}, GP\cite{47}, SSM\cite{19}, RIAD\cite{18}, P-SVDD\cite{28}, and Cutpast\cite{48} as baselines. The corresponding results can be found in \Cref{tab:2}. Overall, the data in Table 2 demonstrates that our method achieved the highest average AUC across all categories, highlighting the effectiveness of our anomaly localization approach.
\begin{table}[htbp]
	\centering
	\caption{Anomaly Localization Performance on the MVTec AD Dataset:AUC Percentages Displayed by Category. The last row shows the average score across all categories. Data for OCGAN, MemAE, AnoGAN, SCADN and SSM is sourced from \cite{19}. MemSTC results are taken from \cite{37} GP results are taken from \cite{47}, RIAD results are referenced from \cite{18}, P-SVDD results are taken from \cite{28} and Cutpast results are taken from \cite{48}. The best-performing methods are highlighted in bold.}
	\resizebox{\linewidth}{!}{
	\begin{tabular}{p{2cm}<{\centering}p{2cm}<{\centering}p{2cm}<{\centering}p{2cm}<{\centering}p{2cm}<{\centering}p{2cm}<{\centering}p{2cm}<{\centering}p{2cm}<{\centering}p{2cm}<{\centering}p{2cm}<{\centering}p{2cm}<{\centering}p{2cm}<{\centering}}
		\toprule
		Category & \textbf{OCGAN} & \textbf{MemAE} & \textbf{AnoGAN} & \textbf{SCADN} & \textbf{MemSTC} & \textbf{GP} & \textbf{SSM} & \textbf{RIAD} & \textbf{P-SVDD} & \textbf{Cutpast} & \cellcolor[rgb]{ .949,  .949,  .949}\textbf{OURS} \\
		 & \cite{35} & \cite{36} & \cite{52} & \cite{38} & \cite{37} & \cite{47} & \cite{19} & \cite{18} & \cite{28} & \cite{48} &   \\
		\midrule
		Bottle & 56.7  & 72.4  & 86    & 69.6  & 87.2  & 93    & 95.9  & \textbf{98.4} & 98.1  & 97.6  & \cellcolor[rgb]{ .949,  .949,  .949}96.6 \\
		Cable & 56.4  & 81.4  & 78    & 81.4  & 91.2  & 94    & 82.1  & 92.8  & \textbf{96.8} & 90    & \cellcolor[rgb]{ .949,  .949,  .949}92.9 \\
		Capsule & 63.7  & 67.3  & 84    & 68.7  & 91.2  & 90    & 98.4  & \textbf{98.8} & 95.8  & 97.4  & \cellcolor[rgb]{ .949,  .949,  .949}96.7 \\
		Carpet & 54.6  & 57.4  & 54    & 64.9  & 85.7  & 96    & 94.4  & \textbf{99.4} & 92.6  & 98.3  & \cellcolor[rgb]{ .949,  .949,  .949}99 \\
		Grid  & 65.2  & 46.8  & 58    & 79.6  & 93.9  & 78    & 99    & 95.7  & 96.2  & 97.5  & \cellcolor[rgb]{ .949,  .949,  .949}\textbf{99.5} \\
		Hazelnut & 84.1  & 84.6  & 87    & 88.4  & 96.1  & 84    & 97.4  & 89.1  & 97.5  & 97.3  & \cellcolor[rgb]{ .949,  .949,  .949}\textbf{99.1} \\
		Leather & 74.9  & 68.6  & 64    & 76.3  & 95.7  & 90    & 99.6  & 87.7  & 97.4  & 99.5  & \cellcolor[rgb]{ .949,  .949,  .949}\textbf{99.5} \\
		MetalNut & 53.4  & 76.9  & 76    & 75.4  & 89    & 91    & 89.6  & 97.8  & \textbf{98} & 93.1  & \cellcolor[rgb]{ .949,  .949,  .949}90.5 \\
		Pill  & 59.6  & 73.7  & 87    & 74.7  & 93.1  & 93    & 97.8  & 84.2  & 95.1  & \textbf{95.7} & \cellcolor[rgb]{ .949,  .949,  .949}93.2 \\
		Screw & 70.8  & 73.2  & 80    & 87.6  & 90.1  & 96    & \textbf{98.9} & 96.3  & 95.7  & 96.7  & \cellcolor[rgb]{ .949,  .949,  .949}98.5 \\
		Tile  & 59.2  & 64.7  & 50    & 67.7  & 85.9  & 80    & 90.2  & \textbf{96.1} & 91.4  & 90.5  & \cellcolor[rgb]{ .949,  .949,  .949}88.5 \\
		Toothbrush & 76.3  & 88.6  & 90    & 90.1  & 95.2  & 96    & 98.9  & 92.5  & 98.1  & 98.1  & \cellcolor[rgb]{ .949,  .949,  .949}\textbf{99} \\
		Transistor & 58.2  & 71.4  & 80    & 68.9  & 86.9  & 100   & 80.1  & 98.8  & \textbf{97} & 93    & \cellcolor[rgb]{ .949,  .949,  .949}92 \\
		Wood  & 65.5  & 65.2  & 62    & 67.2  & 85.1  & 81    & 86.9  & 98.9  & 90.8  & 95.5  & \cellcolor[rgb]{ .949,  .949,  .949}\textbf{96.7} \\
		Zipper & 62.4  & 64.3  & 78    & 67    & 89.4  & 99    & 99    & 85.8  & 95.1  & \textbf{99.3} & \cellcolor[rgb]{ .949,  .949,  .949}99.2 \\
		\midrule
		Mean  & 64.1  & 70.4  & 74    & 75.2  & 90.4  & 91    & 93.9  & 94.2  & 95.7  & 96    & \cellcolor[rgb]{ .949,  .949,  .949}\textbf{96.1} \\
		\bottomrule
	\end{tabular}%
	}
	\label{tab:2}%
\end{table}%
\par
\begin{figure}[!htbp]
	\centering
	\includegraphics[scale=0.18]{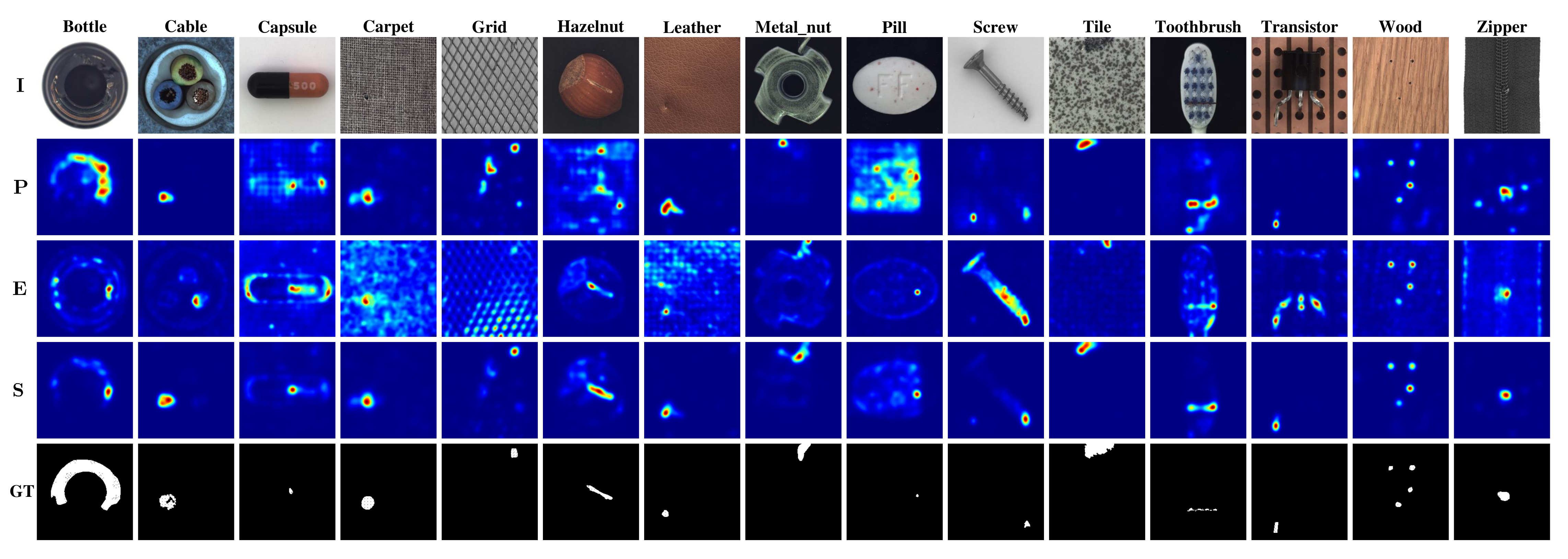}
	\caption{Visualization of our method's anomaly localization for several challenging cases within the MVTec AD dataset. The figure displays the following elements: \textbf{P}: Anomaly score map generated by the classification heade. \textbf{E}: Anomaly score map based on the reconstruction error. \textbf{S}: Final fused anomaly score map. \textbf{GT}: Ground truth label. }
	\label{fig:4}
\end{figure}

\Cref{fig:4} presents the qualitative results of our anomaly localization method for several challenging cases within the MVTec AD dataset. In the figure, \textbf{GT} denotes the ground truth label, \textbf{P} represents the anomaly score map output by the classification head, predicting the likelihood of an anomaly within the input image. \textbf{E} refers to the anomaly score map based on reconstruction error, calculated by comparing the original image with its reconstructed version, and \textbf{S} signifies the final fused anomaly score map, which combines scores from both the classification head and reconstruction error. Our method effectively combines these two approaches, as further demonstrated by the ablation experiments in the subsequent section. These qualitative results highlight how our method accurately pinpoints anomalies in the input image by comparing the ground truth label with the anomaly score maps generated by the classification head and reconstruction error. The final fused anomaly score map (\textbf{S}) provides a more precise localization of the anomaly compared to using P or E alone.
\par
\par
\subsection{Ablation Study}
\label{Ablation experiments}
\begin{figure}[!htbp]
	\centering
	\includegraphics[scale=0.5]{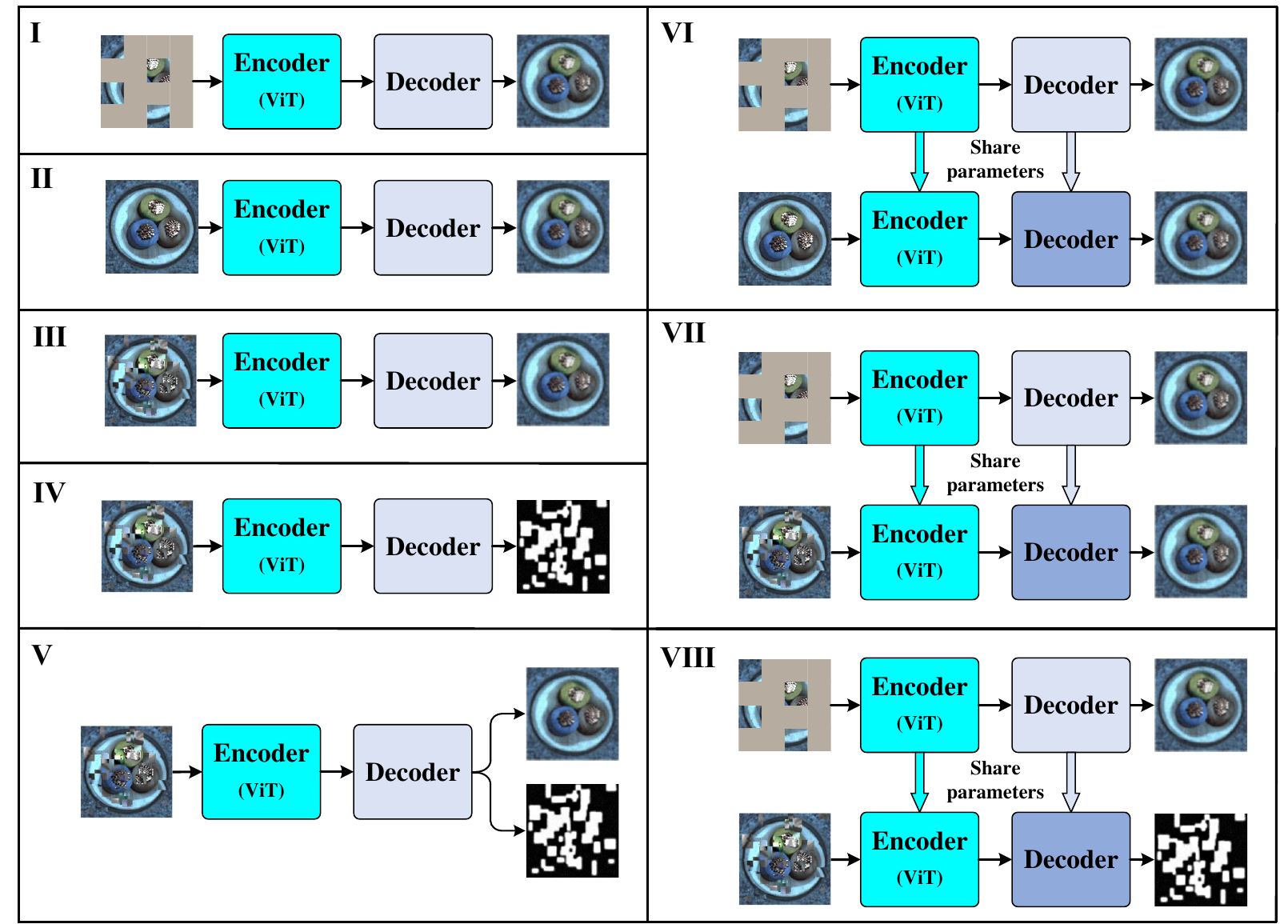}
	\caption{Schematic diagram of the structures for eight Transformer-based anomaly detection models in the ablation experiment. These models serve as both substructures of our approach and independent anomaly detection models. Model I: Applies the MAE model directly for anomaly detection, using reconstruction error determining anomaliess. Model II: Features a basic autoencoder structure, using the entire image as input for both training and detection, with reconstruction error determining anomalies. Model III: Employs a denoising autoencoder, using randomly corrupted normal images as input during training and learning to correct the corruption. Detection is identical to Model II. Model IV: Learns to recognize corrupted regions in randomly corrupted normal images during training. Detection uses pixel classification probability to determine anomaly status and abnormal region. Model V: Similar to our approach but lacks MAE pre-training based on normal images, a feature common to the first five models. Models VI, VII and VIII: Incorporate MAE-based pre-training based on Models II, III and IV, respectively.}
	\label{fig:5}
\end{figure}

\textbf{Experimental Design}: The primary aim of this section is to assess the effectiveness of our proposed model architecture through comprehensive ablation studies. Although Convolutional Neural Networks (CNNs) are predominantly utilized as backbone models in anomaly detection and localization approaches, the integration of Transformers as backbone networks remains relatively unexplored in this field. Consequently, we have devised eight Transformer-based anomaly detection models. These models not only act as substructures of our main model, but also operate as independent anomaly detection models. They employ either reconstruction-based, pixel-classifier based, or a combination of both methodologies for anomaly detection. A detailed description of these models is provided below:
\begin{itemize}
\item Model \uppercase\expandafter{\romannumeral1}: Direct application of the MAE model for anomaly detection, employing a training method identical to MAE, which learns to reconstruct occluded normal images. During detection, an image with 75$\%$ occlusion is input, and the reconstruction error is used to ascertain the anomaly status and identify the abnormal region.
\item Model \uppercase\expandafter{\romannumeral2}: The most basic autoencoder structure, taking the entire image as input during both the training and detection phases, and leveraging reconstruction error to ascertain the anomalous nature of the image.
\item Model \uppercase\expandafter{\romannumeral3}: Denoising autoencoder, in which randomly corrupted normal images are input during training, and the model learns to rectify the corruption. The detection method is identical to Model \uppercase\expandafter{\romannumeral2}.
\item Model \uppercase\expandafter{\romannumeral4}: During training, the model learns to recognize the corrupted regions of randomly corrupted normal images. In detection, the classification probability of pixels is utilized to determine the anomaly status and the abnormal region of the image.
\item Model \uppercase\expandafter{\romannumeral5}: Similar to our approach but lacking MAE pre-training based on normal images. The absence of MAE pre-training is a shared characteristic of the first five models.
\item Model \uppercase\expandafter{\romannumeral6}: The initial stage consists of MAE-based pre-training, while the second stage employs a standard autoencoder structure, with both input and output represented as complete images, utilizing reconstruction error as an anomaly evaluation metric.
\item Models \uppercase\expandafter{\romannumeral7} and \uppercase\expandafter{\romannumeral8}: MAE-based pre-training is incorporated based on Models \uppercase\expandafter{\romannumeral3} and \uppercase\expandafter{\romannumeral4}. On the other hand, Model \uppercase\expandafter{\romannumeral7} is equivalent to our model without pixel classification loss, and Model \uppercase\expandafter{\romannumeral8} is equivalent to our model without reconstruction loss.
\end{itemize}
All training parameters adhere to the descriptions provided in Section \ref{trainsets}. \Cref{fig:5} displays a graphical representation of the eight model structures. \Cref{tab:model} highlights the structural and training method differences among these eight models, as well as the distinctions between them and our proposed model.

\begin{table}[!htbp]
	\centering
	\caption{Comparison of Structural and Training Method Differences Among the Eight Models and Our Proposed Model.}
	\resizebox{\linewidth}{!}{
	\begin{tabular}{m{2cm}<{\centering}|p{3.5cm}<{\centering}  p{4cm}<{\centering} p{4cm}<{\centering} p{4cm}<{\centering} p{4cm}<{\centering}}
		\toprule
		\makecell{\\ \textbf{Model} \\ \\} & \textbf{Two-stage Training} & \textbf{Pixel-level Data Augmentation} & \textbf{Reconstruction Head} & \textbf{Pixel Classifier Head} & \textbf{Fusion Anomaly Matrix} \\
		\midrule
		\textbf{\uppercase\expandafter{\romannumeral1}}     &       &       & \Checkmark     &       &  \\
		\textbf{\uppercase\expandafter{\romannumeral2}}    &       &       & \Checkmark     &       &  \\
		\textbf{\uppercase\expandafter{\romannumeral3}}   &       & \Checkmark     & \Checkmark     &       &  \\
		\textbf{\uppercase\expandafter{\romannumeral4}}    &       & \Checkmark     &       & \Checkmark     &  \\
		\rowcolor[rgb]{ .906,  .902,  .902} \textbf{\uppercase\expandafter{\romannumeral5}}     &       & \Checkmark     & \Checkmark     & \Checkmark     & \Checkmark \\
		\midrule
		\textbf{\uppercase\expandafter{\romannumeral6}}    & \Checkmark     &       & \Checkmark     &       &  \\
		\textbf{\uppercase\expandafter{\romannumeral7}}   & \Checkmark     & \Checkmark     & \Checkmark     &       &  \\
		\textbf{\uppercase\expandafter{\romannumeral8}}  & \Checkmark     & \Checkmark     &       & \Checkmark     &  \\
		\rowcolor[rgb]{ .851,  .851,  .851} \textbf{OURS}  & \Checkmark     & \Checkmark     & \Checkmark     & \Checkmark     & \Checkmark \\
		\bottomrule
	\end{tabular}%
	}
	\label{tab:model}%
\end{table}%

\par
\textbf{Results and Analysis}: \Cref{tab:3} presents a comparison of the anomaly detection outcomes for the eight substructure models and our proposed method, while \Cref{tab:4} contrasts the anomaly localization results. From the experimental findings, the following insights can be derived:
\begin{table}[htbp]
	\centering
	\caption{Ablation Study Results for Abnormal Detection: AUC Percentages Displayed. Model I is the original MAE model. Model II is a basic Transformer-based autoencoder. Model III is a denoising autoencoder, and Model IV is a self-supervised pixel classification model. Model V is a fusion model combining Model III and Model IV. Models VI through VIII are modifications of Models II to IV, incorporating MAE pre-training on normal data. The final column presents results from our method, which is identical to Model V but includes MAE pre-training on normal data. We also compared these models to the basic CNN-based autoencoder anomaly detection method, DAE, using results from\cite{19}. The best-performing method is emphasized in bold.}
	\resizebox{\linewidth}{!}{
			\begin{tabular}{p{2cm}<{\centering}|p{2cm}<{\centering}|p{2cm}<{\centering}p{2cm}<{\centering}p{2cm}<{\centering}p{2cm}<{\centering}|p{2cm}<{\centering}|p{2cm}<{\centering}p{2cm}<{\centering}p{2cm}<{\centering}|p{2cm}<{\centering}}
			\toprule
			Category & 
			\textbf{DAE}\cite{15} & 
			\textbf{Model \uppercase\expandafter{\romannumeral1}} & \textbf{\uppercase\expandafter{\romannumeral2}} & \textbf{\uppercase\expandafter{\romannumeral3}} & \textbf{\uppercase\expandafter{\romannumeral4}} & \textbf{\uppercase\expandafter{\romannumeral5}} & \textbf{\uppercase\expandafter{\romannumeral6}} & \textbf{\uppercase\expandafter{\romannumeral7}} & \textbf{\uppercase\expandafter{\romannumeral8}} & \cellcolor[rgb]{ .949,  .949,  .949}\textbf{OURS} \\
			\midrule
			bottle & 86.0 & 85.3  & 71.4  & 89.0  & 83.7  & 92.9  & 83.3  & 96.1  & 100.0  & \cellcolor[rgb]{ .949,  .949,  .949}\textbf{100.0 } \\
			cable & 64.8 & 77.8  & 61.4  & 78.7  & 60.5  & 79.3  & 52.8  & 68.3  & 93.4  & \cellcolor[rgb]{ .949,  .949,  .949}\textbf{93.6 } \\
			capsule & 53.4 & 67.1  & 81.2  & 72.8  & 74.8  & 79.8  & 79.2  & 82.6  & 89.2  &\cellcolor[rgb]{ .949,  .949,  .949} \textbf{92.9 } \\
			carpet & 58.8 & 46.2  & 46.7  & 51.2  & 84.5  & 73.2  & 40.9  & 46.1  & \textbf{98.9 } & \cellcolor[rgb]{ .949,  .949,  .949}97.1  \\
			grid  & 85.8 & 94.0  & 82.1  & 84.8  & 96.2  & 96.8  & 86.0  & 86.9  & 100.0  & \cellcolor[rgb]{ .949,  .949,  .949}\textbf{100.0 } \\
			hazelnut & 51.3 & 90.4  & 96.6  & 90.3  & 89.7  & 90.9  & 93.7  & 88.1  & 99.9  & \cellcolor[rgb]{ .949,  .949,  .949}\textbf{100.0 } \\
			leather & 49.7 & 52.8  & 44.4  & 54.1  & 99.4  & 96.0  & 47.4  & 57.2  & 100.0  & \cellcolor[rgb]{ .949,  .949,  .949}\textbf{100.0 } \\
			metal\_nut & 79.3 & 56.5  & 47.3  & 63.6  & 92.6  & 94.7  & 29.0  & 56.6  & \textbf{99.9 } & \cellcolor[rgb]{ .949,  .949,  .949}99.3  \\
			pill  & 69.3 & 59.0  & 86.3  & 76.8  & 82.8  & 84.5  & 89.4  & 88.1  & \textbf{97.4 } & \cellcolor[rgb]{ .949,  .949,  .949}96.6  \\
			screw & 71.9 & 100.0  & 99.8  & 89.8  & \textbf{100.0 } & 98.8  & 99.7  & 99.9  & 82.9  & \cellcolor[rgb]{ .949,  .949,  .949}88.3  \\
			tile  & 89.4 & 75.6  & 92.5  & 75.9  & 90.6  & 95.7  & 87.3  & 92.2  & 100.0  & \cellcolor[rgb]{ .949,  .949,  .949}\textbf{100.0 } \\
			toothbrush & 94.2 & 80.6  & 89.7  & 98.1  & 94.2  & 98.3  & 90.8  & 97.8  & 100.0  & \cellcolor[rgb]{ .949,  .949,  .949}\textbf{100.0 } \\
			transistor & 37.6 & 84.3  & 71.4  & 76.0  & 87.2  & 89.7  & 72.5  & 83.4  & \textbf{97.2 } & \cellcolor[rgb]{ .949,  .949,  .949}96.6  \\
			wood  & 88.2 & 95.0  & 96.4  & 95.0  & 99.7  & 99.5  & 92.0  & 94.0  & 100.0  &\cellcolor[rgb]{ .949,  .949,  .949}\textbf{100.0 } \\
			zipper & 81.9 & 75.8  & 78.5  & 75.9  & 93.4  & 92.5  & 84.2  & 90.3  & 100.0  & \cellcolor[rgb]{ .949,  .949,  .949}\textbf{100.0 } \\
			\midrule
			mean  & 70.7 & 76.0  & 76.4  & 78.1  & 88.6  & 90.8  & 75.2  & 81.8  & 97.3  & \cellcolor[rgb]{ .949,  .949,  .949}\textbf{97.6 } \\
			\bottomrule
		\end{tabular}%
	}
	\label{tab:3}%
\end{table}%

\begin{table}[htbp]
	\centering
	\caption{Ablation Study Results for Abnormal Localization with AUC Percentages, Including DAE from \cite{19} (Best-Performing Method in Bold)}
	\resizebox{\linewidth}{!}{
	\begin{tabular}{p{2cm}<{\centering}|p{2cm}<{\centering}|p{2cm}<{\centering}p{2cm}<{\centering}p{2cm}<{\centering}p{2cm}<{\centering}|p{2cm}<{\centering}|p{2cm}<{\centering}p{2cm}<{\centering}p{2cm}<{\centering}|p{2cm}<{\centering}}
		\toprule
		Category & \textbf{DAE}\cite{15}& \textbf{Model \uppercase\expandafter{\romannumeral1}} & \textbf{\uppercase\expandafter{\romannumeral2}} & \textbf{\uppercase\expandafter{\romannumeral3}} & \textbf{\uppercase\expandafter{\romannumeral4}} & \textbf{\uppercase\expandafter{\romannumeral5}} & \textbf{\uppercase\expandafter{\romannumeral6}} & \textbf{\uppercase\expandafter{\romannumeral7}} & \textbf{\uppercase\expandafter{\romannumeral8}} & \cellcolor[rgb]{ .949,  .949,  .949}\textbf{OURS} \\
		\midrule
		bottle & 54.4 & 74.3  & 79.1  & 88.9  & 79.0  & 91.4  & 84.2  & 92.2  & 92.7  & \cellcolor[rgb]{ .949,  .949,  .949}\textbf{96.6 } \\
		cable & 53.5 & 91.6  & 85.1  & 89.5  & 79.4  & 89.8  & 83.7  & 86.9  & 87.0  & \cellcolor[rgb]{ .949,  .949,  .949}\textbf{92.9 } \\
		capsule & 54.2 & 88.1  & 94.1  & 95.1  & 84.2  & 94.4  & 93.3  & 95.5  & 93.6  & \cellcolor[rgb]{ .949,  .949,  .949}\textbf{96.7 } \\
		carpet & 52.8 & 82.2  & 81.4  & 73.5  & 74.0  & 70.3  & 77.6  & 82.4  & 98.9  & \cellcolor[rgb]{ .949,  .949,  .949}\textbf{99.0 } \\
		grid  & 55.0 & 99.2  & 79.0  & 70.6  & 74.7  & 79.2  & 94.7  & 98.2  & 99.3  & \cellcolor[rgb]{ .949,  .949,  .949}\textbf{99.5 } \\
		hazelnut & 66.4 & 97.1  & 98.2  & 98.3  & 82.4  & 96.2  & 98.2  & 98.3  & 97.7  & \cellcolor[rgb]{ .949,  .949,  .949}\textbf{99.1 } \\
		leather & 78.3 & 97.5  & 95.6  & 97.1  & 83.5  & 95.6  & 88.5  & 96.5  & 99.3  & \cellcolor[rgb]{ .949,  .949,  .949}\textbf{99.5 } \\
		metal\_nut & 53.9 & 88.3  & 70.5  & \textbf{90.9 } & 67.2  & 78.3  & 71.2  & 89.3  & 67.0  & \cellcolor[rgb]{ .949,  .949,  .949}90.5  \\
		pill  & 55.5 & 56.6  & 96.7  & 92.2  & 74.9  & 89.9  & 97.7  & \textbf{97.8 } & 85.5  & \cellcolor[rgb]{ .949,  .949,  .949}93.2  \\
		screw & 57.0 & 97.9  & 97.7  & 97.4  & 73.0  & 96.3  & 98.2  & \textbf{99.2 } & 96.4  & \cellcolor[rgb]{ .949,  .949,  .949}98.5  \\
		tile  & 63.0 & 60.3  & 89.6  & 61.6  & 67.4  & 69.3  & 77.2  & 83.9  & 84.7  & \cellcolor[rgb]{ .949,  .949,  .949}\textbf{88.5 } \\
		toothbrush & 61.6 & 70.4  & 96.3  & 98.2  & 79.4  & 97.6  & 95.0  & 98.4  & 98.1  & \cellcolor[rgb]{ .949,  .949,  .949}\textbf{99.0 } \\
		transistor & 53.2 & 88.0  & 76.7  & 87.3  & 89.1  & \textbf{94.0 } & 77.6  & 86.3  & 94.7  & \cellcolor[rgb]{ .949,  .949,  .949}92.0  \\
		wood  & 61.2 & 85.2  & 86.7  & 87.8  & 78.5  & 95.0  & 82.2  & 87.1  & 95.3  & \cellcolor[rgb]{ .949,  .949,  .949}\textbf{96.7 } \\
		zipper & 53.6 & 86.0  & 86.8  & 88.2  & 72.7  & 89.2  & 92.7  & 92.7  & 98.7  & \cellcolor[rgb]{ .949,  .949,  .949}\textbf{99.2 } \\
		\midrule
		mean  & 58.2 & 84.2  & 87.6  & 87.8  & 77.3  & 88.4  & 87.5  & 92.3  & 92.6  & \cellcolor[rgb]{ .949,  .949,  .949}\textbf{96.1 } \\
		\bottomrule
	\end{tabular}%
	}
	\label{tab:4}%
\end{table}%

\begin{itemize}
\item Models that undergo MAE pre-training on normal data, as seen in Models \uppercase\expandafter{\romannumeral7} and \uppercase\expandafter{\romannumeral8}, perform notably better than models without pretraining, such as Models \uppercase\expandafter{\romannumeral3} and \uppercase\expandafter{\romannumeral4}. This is because the pre-training stage provides the model with distribution information about normal samples, allowing it to express this distribution better during the anomaly training phase. This method can be considered a form of incremental learning, where the model's performance is enhanced by introducing new tasks and data (in this case, new image enhancement techniques). This also explains why pre-training did not boost Model \uppercase\expandafter{\romannumeral6}'s performance compared to Model \uppercase\expandafter{\romannumeral2}, as both Model \uppercase\expandafter{\romannumeral6}'s training stages fundamentally employ autoencoders without introducing new tasks and data.

\item Self-supervised pixel classification-based approaches demonstrate significantly higher anomaly detection performance than reconstruction-based methods, as evidenced by the comparison between Models \uppercase\expandafter{\romannumeral7} and \uppercase\expandafter{\romannumeral8}. These two models employ consistent image enhancement techniques and share the same pre-training model parameters. This indicates that our proposed self-supervised pixel classification-based anomaly detection approach, which utilizes features from each decoder's Transformer block (from high-level semantic, global features to low-level texture, and partial features), substantially improves anomaly detection performance compared to reconstruction methods.

\item In the absence of pre-training, reconstruction-based methods perform better in anomaly localization than self-supervised pixel classification-based methods, as evidenced by the comparison between Models \uppercase\expandafter{\romannumeral3} and \uppercase\expandafter{\romannumeral4}. However, when pre-training is incorporated, the performance differences between these two methods in anomaly localization become minimal, as observed in Models \uppercase\expandafter{\romannumeral7} and \uppercase\expandafter{\romannumeral8}. This is because during the anomaly training phase, we fix the encoder parameters. For end-to-end pixel classifiers, the encoder-decoder structure is overly redundant during training. By fixing the encoder parameters, this redundancy is eliminated, enhancing the model's learning capability.

\item Model \uppercase\expandafter{\romannumeral2} features the most basic autoencoder structure, sharing the same anomaly detection principle as DAE, which detects anomalies via reconstruction errors. The only difference is that DAE relies on CNNs, while Model \uppercase\expandafter{\romannumeral2} uses Transformers. The experimental results indicate that Model II outperforms DAE, particularly in anomaly localization. We believe this is because Transformers are more proficient at learning continuous features between pixels compared to CNNs, and in the MVTec AD dataset, defect anomalies often appear as discontinuities between pixels.

\item Our fusion model achieves the highest AUC metrics in both anomaly detection and localization, demonstrating that it effectively capitalizes on the advantages of both reconstruction and self-supervised pixel classification-based approaches.
\end{itemize}

\section{Conclusion}
\label{}
In this paper, we introduce an incremental self-supervised learning approach for anomaly detection and localization, employing the Transformer as the backbone network. Our model comprises a ViT encoder, a Transformer-based decoder, a reconstruction head, and a multi-level feature pixel classification head. We adopt a two-stage incremental learning training strategy: during the first stage, we enhance the encoder's feature extraction capabilities using the Masked Autoencoder method. In the second stage, we implement an innovative pixel-level self-supervised learning method, merging reconstruction and pixel classification to refine the decoder's capacity for anomaly detection and localization.
\par
Our ablation experiments demonstrate that this two-stage training approach significantly boosts the model's anomaly detection performance. Moreover, our fusion model effectively integrates the strengths of reconstruction and pixel classification methods, resulting in improved detection accuracy. Our experiments on the MVTec AD dataset further indicate that our method surpasses some of the leading anomaly detection techniques that employ CNNs as the backbone network.







\bibliographystyle{plain}
\bibliography{my}  

\begin{thebibliography}{10}

\bibitem{33}
Samet Akcay, Amir Atapour-Abarghouei, and Toby~P. Breckon.
\newblock {GANomaly}: Semi-supervised anomaly detection via adversarial
  training.
\newblock In {\em Computer Vision {\textendash} {ACCV} 2018}, pages 622--637,
  2019.

\bibitem{34}
Samet Akcay, Amir Atapour-Abarghouei, and Toby~P. Breckon.
\newblock Skip-{GANomaly}: Skip connected and adversarially trained
  encoder-decoder anomaly detection.
\newblock In {\em 2019 International Joint Conference on Neural Networks
  ({IJCNN})}. {IEEE}, July 2019.

\bibitem{7}
Samet Akcay and Toby Breckon.
\newblock Towards automatic threat detection: A survey of advances of deep
  learning within x-ray security imaging.
\newblock {\em Pattern Recognition}, 122:108245, 2022.

\bibitem{24}
Eden Belouadah, Adrian Popescu, and Ioannis Kanellos.
\newblock A comprehensive study of class incremental learning algorithms for
  visual tasks.
\newblock {\em Neural Networks}, 135:38--54, 2021.

\bibitem{46}
Paul Bergmann, Michael Fauser, David Sattlegger, and Carsten Steger.
\newblock {MVTec} {AD} {\textemdash} a comprehensive real-world dataset for
  unsupervised anomaly detection.
\newblock In {\em 2019 {IEEE}/{CVF} Conference on Computer Vision and Pattern
  Recognition ({CVPR})}. {IEEE}, June 2019.

\bibitem{6}
Raghavendra Chalapathy and Sanjay Chawla.
\newblock Deep learning for anomaly detection: A survey.
\newblock {\em arXiv preprint arXiv:1901.03407}, 2019.

\bibitem{1}
Varun Chandola, Arindam Banerjee, and Vipin Kumar.
\newblock Anomaly detection.
\newblock {\em {ACM} Computing Surveys}, 41(3):1--58, July 2009.

\bibitem{53}
Liyang Chen, Zhiyuan You, Nian Zhang, Juntong Xi, and Xinyi Le.
\newblock Utrad: Anomaly detection and localization with u-transformer.
\newblock {\em Neural Networks}, 147:53--62, 2022.

\bibitem{41}
X.~Chen, M.~Ding, X.~Wang, Y.~Xin, S.~Mo, Y.~Wang, S.~Han, P.~Luo, G.~Zeng, and
  J.~Wang.
\newblock Context autoencoder for self-supervised representation learning.
\newblock 2022.

\bibitem{25}
Corinna Cortes and Vladimir Vapnik.
\newblock {\em Machine Learning}, 20(3):273--297, 1995.

\bibitem{9}
Y.~Cui, Z.~Liu, and S.~Lian.
\newblock A survey on unsupervised visual industrial anomaly detection
  algorithms.
\newblock {\em arXiv e-prints}, 2022.

\bibitem{14}
Tal Daniel, Thanard Kurutach, and Aviv Tamar.
\newblock Deep variational semi-supervised novelty detection.
\newblock {\em arXiv preprint arXiv:1911.04971}, 2019.

\bibitem{21}
Alexey Dosovitskiy, Lucas Beyer, Alexander Kolesnikov, Dirk Weissenborn,
  Xiaohua Zhai, Thomas Unterthiner, Mostafa Dehghani, Matthias Minderer, Georg
  Heigold, Sylvain Gelly, et~al.
\newblock An image is worth 16x16 words: Transformers for image recognition at
  scale.
\newblock {\em arXiv preprint arXiv:2010.11929}, 2020.

\bibitem{4}
Eleazar Eskin.
\newblock Anomaly detection over noisy data using learned probability
  distributions.
\newblock In {\em The Seventeenth International Conference on Machine Learning
  (ICML-2000)}, 2000.

\bibitem{42}
Christoph Feichtenhofer, Haoqi Fan, Yanghao Li, and Kaiming He.
\newblock Masked autoencoders as spatiotemporal learners, 2022.

\bibitem{12}
Izhak Golan and Ran El-Yaniv.
\newblock Deep anomaly detection using geometric transformations.
\newblock In S.~Bengio, H.~Wallach, H.~Larochelle, K.~Grauman, N.~Cesa-Bianchi,
  and R.~Garnett, editors, {\em Advances in Neural Information Processing
  Systems}, volume~31. Curran Associates, Inc., 2018.

\bibitem{36}
Dong Gong, Lingqiao Liu, Vuong Le, Budhaditya Saha, Moussa~Reda Mansour, Svetha
  Venkatesh, and Anton Van~Den Hengel.
\newblock Memorizing normality to detect anomaly: Memory-augmented deep
  autoencoder for unsupervised anomaly detection.
\newblock In {\em 2019 {IEEE}/{CVF} International Conference on Computer Vision
  ({ICCV})}. {IEEE}, October 2019.

\bibitem{20}
Kaiming He, Xinlei Chen, Saining Xie, Yanghao Li, Piotr Doll\'ar, and Ross
  Girshick.
\newblock Masked autoencoders are scalable vision learners.
\newblock In {\em Proceedings of the IEEE/CVF Conference on Computer Vision and
  Pattern Recognition (CVPR)}, pages 16000--16009, June 2022.

\bibitem{19}
Chaoqin Huang, Qinwei Xu, Yanfeng Wang, Yu~Wang, and Ya~Zhang.
\newblock Self-supervised masking for unsupervised anomaly detection and
  localization.
\newblock {\em IEEE Transactions on Multimedia}, pages 1--1, 2022.

\bibitem{40}
Mike Huisman, Jan~N. van Rijn, and Aske Plaat.
\newblock A survey of deep meta-learning.
\newblock {\em Artificial Intelligence Review}, 54(6):4483--4541, April 2021.

\bibitem{44}
A.~Krizhevsky and G.~Hinton.
\newblock Learning multiple layers of features from tiny images.
\newblock {\em Handbook of Systemic Autoimmune Diseases}, 1(4), 2009.

\bibitem{48}
Chun-Liang Li, Kihyuk Sohn, Jinsung Yoon, and Tomas Pfister.
\newblock {CutPaste}: Self-supervised learning for anomaly detection and
  localization.
\newblock In {\em 2021 {IEEE}/{CVF} Conference on Computer Vision and Pattern
  Recognition ({CVPR})}. {IEEE}, June 2021.

\bibitem{51}
Jonathan Long, Evan Shelhamer, and Trevor Darrell.
\newblock Fully convolutional networks for semantic segmentation.
\newblock In {\em 2015 {IEEE} Conference on Computer Vision and Pattern
  Recognition ({CVPR})}. {IEEE}, June 2015.

\bibitem{13}
Poojan Oza and Vishal~M Patel.
\newblock Deep cnn-based multi-task learning for open-set recognition.
\newblock {\em arXiv preprint arXiv:1903.03161}, 2019.

\bibitem{35}
Pramuditha Perera, Ramesh Nallapati, and Bing Xiang.
\newblock {OCGAN}: One-class novelty detection using {GANs} with constrained
  latent representations.
\newblock In {\em 2019 {IEEE}/{CVF} Conference on Computer Vision and Pattern
  Recognition ({CVPR})}. {IEEE}, June 2019.

\bibitem{11}
Esteban Reyes and Pablo~A. Estévez.
\newblock Transformation based deep anomaly detection in astronomical images.
\newblock In {\em 2020 International Joint Conference on Neural Networks
  (IJCNN)}, pages 1--8, 2020.

\bibitem{27}
L.~Ruff, R.~A. Vandermeulen, N~Görnitz, L.~Deecke, and M.~Kloft.
\newblock Deep one-class classification.
\newblock In {\em International Conference on Machine Learning}, 2018.

\bibitem{15}
Mayu Sakurada and Takehisa Yairi.
\newblock Anomaly detection using autoencoders with nonlinear dimensionality
  reduction.
\newblock In {\em Proceedings of the MLSDA 2014 2nd Workshop on Machine
  Learning for Sensory Data Analysis}, MLSDA'14, pages 4--11, New York, NY,
  USA, 2014.

\bibitem{52}
Thomas Schlegl, Philipp Seeb\"{o}ck, Sebastian~M. Waldstein, Ursula
  Schmidt-Erfurth, and Georg Langs.
\newblock Unsupervised anomaly detection with generative adversarial networks
  to guide marker discovery.
\newblock In {\em Lecture Notes in Computer Science}, pages 146--157, 2017.

\bibitem{5}
Bernhard Sch\"{o}lkopf, John~C. Platt, John Shawe-Taylor, Alex~J. Smola, and
  Robert~C. Williamson.
\newblock Estimating the support of a high-dimensional distribution.
\newblock Technical report, Technical Report MSR-T R-99--87, Microsoft Research
  (MSR), 1999.

\bibitem{26}
David~M.J. Tax and Robert~P.W. Duin.
\newblock Support vector data description.
\newblock {\em Machine Learning}, 54(1):45--66, January 2004.

\bibitem{23}
Yu~Tian, Guansong Pang, Yuyuan Liu, Chong Wang, Yuanhong Chen, Fengbei Liu,
  Rajvinder Singh, Johan~W Verjans, and Gustavo Carneiro.
\newblock Unsupervised anomaly detection in medical images with a
  memory-augmented multi-level cross-attentional masked autoencoder.
\newblock {\em arXiv preprint arXiv:2203.11725}, 2022.

\bibitem{8}
Maximilian~E. Tschuchnig and Michael Gadermayr.
\newblock Anomaly detection in medical imaging - a mini review.
\newblock In Peter Haber, Thomas~J. Lampoltshammer, Helmut Leopold, and Manfred
  Mayr, editors, {\em Data Science -- Analytics and Applications}, pages
  33--38, Wiesbaden, 2022.

\bibitem{22}
Ashish Vaswani, Noam Shazeer, Niki Parmar, Jakob Uszkoreit, Llion Jones,
  Aidan~N Gomez, \L~ukasz Kaiser, and Illia Polosukhin.
\newblock Attention is all you need.
\newblock In I.~Guyon, U.~Von Luxburg, S.~Bengio, H.~Wallach, R.~Fergus,
  S.~Vishwanathan, and R.~Garnett, editors, {\em Advances in Neural Information
  Processing Systems}, volume~30. Curran Associates, Inc., 2017.

\bibitem{47}
Shenzhi Wang, Liwei Wu, Lei Cui, and Yujun Shen.
\newblock Glancing at the patch: Anomaly localization with global and local
  feature comparison.
\newblock In {\em 2021 {IEEE}/{CVF} Conference on Computer Vision and Pattern
  Recognition ({CVPR})}. {IEEE}, June 2021.

\bibitem{49}
Z.~Wang, A.C. Bovik, H.R. Sheikh, and E.P. Simoncelli.
\newblock Image quality assessment: From error visibility to structural
  similarity.
\newblock {\em {IEEE} Transactions on Image Processing}, 13(4):600--612, April
  2004.

\bibitem{43}
Chen Wei, Haoqi Fan, Saining Xie, Chao-Yuan Wu, Alan Yuille, and Christoph
  Feichtenhofer.
\newblock Masked feature prediction for self-supervised visual pre-training.
\newblock In {\em 2022 {IEEE}/{CVF} Conference on Computer Vision and Pattern
  Recognition ({CVPR})}. {IEEE}, June 2022.

\bibitem{39}
Jhih-Ciang Wu, Ding-Jie Chen, Chiou-Shann Fuh, and Tyng-Luh Liu.
\newblock Learning unsupervised metaformer for anomaly detection.
\newblock In {\em 2021 {IEEE}/{CVF} International Conference on Computer Vision
  ({ICCV})}. {IEEE}, October 2021.

\bibitem{2}
Xuan Xia, Xizhou Pan, Nan Li, Xing He, Lin Ma, Xiaoguang Zhang, and Ning Ding.
\newblock Gan-based anomaly detection: A review.
\newblock {\em Neurocomputing}, 493:497--535, 2022.

\bibitem{45}
H.~Xiao, K.~Rasul, and R.~Vollgraf.
\newblock Fashion-mnist: a novel image dataset for benchmarking machine
  learning algorithms.
\newblock 2017.

\bibitem{3}
Kenji Yamanishi, Jun-Ichi Takeuchi, Graham Williams, and Peter Milne.
\newblock On-line unsupervised outlier detection using finite mixtures with
  discounting learning algorithms.
\newblock {\em Data Min. Knowl. Discov.}, 8(3):275--300, may 2004.

\bibitem{38}
Xudong Yan, Huaidong Zhang, Xuemiao Xu, Xiaowei Hu, and Pheng-Ann Heng.
\newblock Learning semantic context from normal samples for unsupervised
  anomaly detection.
\newblock {\em Proceedings of the {AAAI} Conference on Artificial
  Intelligence}, 35(4):3110--3118, May 2021.

\bibitem{28}
Jihun Yi and Sungroh Yoon.
\newblock Patch {SVDD}: Patch-level {SVDD} for anomaly detection and
  segmentation.
\newblock In {\em Computer Vision {\textendash} {ACCV} 2020}, pages 375--390,
  2021.

\bibitem{18}
Vitjan Zavrtanik, Matej Kristan, and Danijel Skočaj.
\newblock Reconstruction by inpainting for visual anomaly detection.
\newblock {\em Pattern Recognition}, 112:107706, 2021.

\bibitem{37}
Kang Zhou, Jing Li, Yuting Xiao, Jianlong Yang, Jun Cheng, Wen Liu, Weixin Luo,
  Jiang Liu, and Shenghua Gao.
\newblock Memorizing structure-texture correspondence for image anomaly
  detection.
\newblock {\em {IEEE} Transactions on Neural Networks and Learning Systems},
  33(6):2335--2349, June 2022.

\end{thebibliography}

\end{document}